\documentclass{article} 

\usepackage{microtype}
\usepackage{graphicx}
\usepackage{subcaption}
\usepackage{booktabs} 
\usepackage{amsmath,amssymb,amsfonts,amsthm}
\usepackage{mathtools}
\usepackage{algorithmic}
\usepackage{algorithm}
\usepackage{graphicx}
\usepackage{textcomp}
\usepackage{xcolor}
\usepackage[table]{xcolor}
\usepackage{color,colortbl}
\usepackage{wrapfig}
\usepackage{array}
\usepackage{bbding}
\usepackage{arydshln}
\usepackage{amsmath}
\usepackage{amssymb}
\usepackage{graphicx}
\usepackage{CJKutf8}
\usepackage{multirow}
\usepackage{enumitem} 
\usepackage{makecell}
\usepackage{bbm}
\usepackage{arydshln}
\usepackage{hyperref}
\usepackage{tcolorbox}
\tcbuselibrary{breakable, skins}

\makeatletter
\newenvironment{breakablealgorithm}
{%
    \par\addvspace{\intextsep}%
    \refstepcounter{algorithm}%
    \noindent\hrule height 0.8pt depth 0pt\kern 2pt%
    \renewcommand{\caption}[2][\relax]{%
        {\raggedright\textbf{\ALG@name~\thealgorithm:} ##2\par}%
        \ifx\relax##1\relax
            \addcontentsline{loa}{algorithm}{\protect\numberline{\thealgorithm}##2}%
        \else
            \addcontentsline{loa}{algorithm}{\protect\numberline{\thealgorithm}##1}%
        \fi
        \kern 2pt\hrule\kern 2pt%
    }%
}
{%
    \kern 2pt\hrule\par\addvspace{\intextsep}%
}
\makeatother

\newcolumntype{I}{!{\vrule width 1.2pt}}
\newlength\savedwidth
\newcommand\whline{\noalign{\global\savedwidth\arrayrulewidth
                            \global\arrayrulewidth 1.5pt}
                   \hline
                   \noalign{\global\arrayrulewidth\savedwidth}}
\newlength\savewidth

\newcommand{\s}[1]{}
\definecolor{darkergreen}{RGB}{21, 152, 56}
\newcommand\greenp[1]{\textcolor{darkergreen}{}}

\let\CoCAAddContentsLine\addcontentsline
\usepackage{iclr2027_conference}
\let\addcontentsline\CoCAAddContentsLine
\usepackage{qwen_character}

\definecolor{newcolor}{rgb}{.8,.349,.1}
\definecolor{mypink}{rgb}{.99,.91,.95}
\definecolor{mygreen}{HTML}{19A576}
\definecolor{lgreen}{HTML}{00B050}
\definecolor{mygray}{HTML}{eeeeee}
\definecolor{myred}{HTML}{fae4df}
\definecolor{mydarkred}{HTML}{ed3333}
\definecolor{mydarkgray}{HTML}{808080}
\definecolor{myblue}{HTML}{E0F0FA}
\definecolor{mypink}{rgb}{0,0,0}
\definecolor{myyellow}{HTML}{FCF0C7}
\definecolor{mypurple}{HTML}{E3E2FC}
\definecolor{cocapurple}{HTML}{615CED}

\usepackage{amsmath,amsfonts,bm}

\def\eqref#1{equation~\ref{#1}}

\def\1{\bm{1}}

\DeclareMathAlphabet{\mathsfit}{\encodingdefault}{\sfdefault}{m}{sl}
\SetMathAlphabet{\mathsfit}{bold}{\encodingdefault}{\sfdefault}{bx}{n}

\usepackage{hyperref}
\usepackage{url}

\title{Co-Evolving Policy and Rubric via Capability Awareness for Open-Ended Generation
}

\author{%
  \textbf{Chengsheng Zhang\textsuperscript{1,2,}\thanks{Work done during research internship at Alibaba Token Hub.} \quad
  Wenshen Xu\textsuperscript{2} \quad
  Yuchuan Wu\textsuperscript{2} \quad
  Zhiyuan Yu\textsuperscript{2,3}} \\
  \textbf{Junfeng Fang\textsuperscript{4} \quad
  Tianlong Wang\textsuperscript{2} \quad
  Bingxue Qiu\textsuperscript{2} \quad
  Yongbin Li\textsuperscript{2,}\thanks{Corresponding authors: \texttt{shuide.lyb@alibaba-inc.com}, \texttt{xiangwang1123@gmail.com}.} \quad
  Xiang Wang\textsuperscript{1,}\footnotemark[2]} \\[4pt]
  \textsuperscript{1}University of Science and Technology of China \quad
  \textsuperscript{2}Alibaba Token Hub, Alibaba Group \\
  \textsuperscript{3}Nanjing University \quad
  \textsuperscript{4}National University of Singapore \\[4pt]
  \texttt{zhangcs66@mail.ustc.edu.cn}
}

\iclrfinalcopy 
\begin{document}

\begin{NoHyper}
\maketitle
\end{NoHyper}

\begin{abstract}

Rubric-based reinforcement learning (RL) offers interpretable, fine-grained supervision for aligning large language models (LLMs)  on open-ended generation tasks.
However, static rubric criteria may become saturated or remain beyond the policy's exploration, rendering the reward signals insufficiently discriminative.
While recent efforts seek to guide sampling exploration or evolve rubrics, 
they fail to explicitly factor in how rubric difficulty should track the evolving policy's capabilities, resulting in a persistent mismatch between the optimization objective and the policy's current capability frontier.
In this paper, we propose \textbf{Co}-Evolving Policy and Rubric via \textbf{C}apability \textbf{A}wareness (\textbf{CoCA}), a closed-loop framework that jointly adapts the rubric and policy through criterion-level diagnosis.
Specifically, CoCA contrasts query-only and rubric-guided rollouts to characterize each criterion relative to the current policy, recalibrates criteria with mismatched difficulty, and identifies guidance-revealed capability gaps for frontier-based learning.
Based on this insight, CoCA either expands rubric-conditioned frontiers through guided optimization or transfers qualified gains through on-policy self-distillation.
As the policy updates, CoCA improves the capability diagnosis and enables more reliable distillation, thereby completing the co-evolution loop.
Extensive experiments across diverse benchmarks and LLMs demonstrate that CoCA consistently improves performance over rubric-based RL methods.
Moreover, CoCA promotes the internalization of guidance-revealed capabilities into query-only behavior while mitigating capability degradation.

\end{abstract}

\section{Introduction}
\label{sec:introduction}

\begin{figure*}[!th]
    \centering
    \begin{minipage}[t]{\linewidth}
        \centering
        \IfFileExists{figures/intro.pdf}{%
            \includegraphics[width=\linewidth]{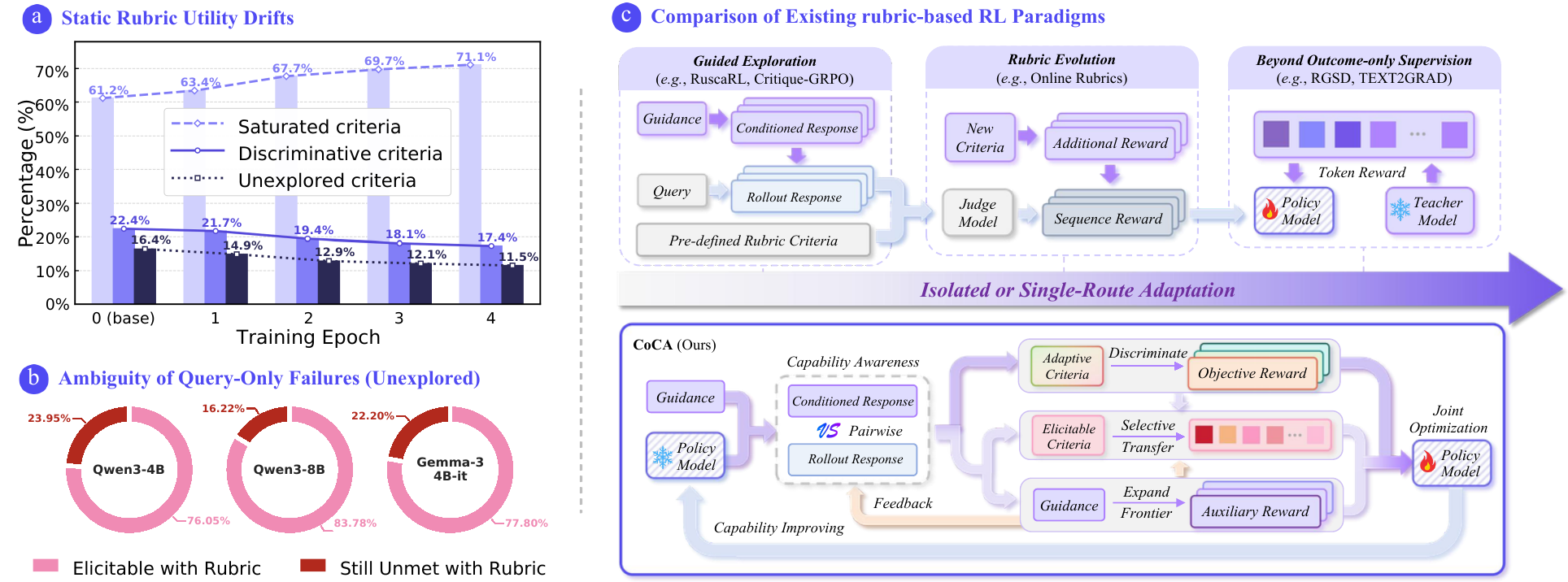}%
        }{%
            \fbox{%
                \parbox[c][3.1cm][c]{0.96\linewidth}{%
                    \centering
                    \textbf{Figure 1 placeholder}\par
                    Static-rubric trajectory $\rightarrow$ query-only failure decomposition
                    $\rightarrow$ diagnosis-routed co-evolution
                }%
            }%
        }
    \end{minipage}
    \caption{
        Motivation and overview of CoCA. 
        \textbf{(a)} Static criteria may become saturated or remain unmet, limiting reward discriminativeness.
        \textbf{(b)} Query-only failures can conceal capabilities elicitable under rubric guidance.
        \textbf{(c)} Distinct from existing rubric-based RL methods, CoCA tracks the policy's capability boundary to couple rubric adaptation and policy learning in a closed loop.
    }
    \label{imgs:intro}
\end{figure*}

Large language models (LLMs) have demonstrated remarkable capabilities in open-ended generation, signaling a shift from functional tools toward personified partners~\citep{arora2025healthbench,liu2025heartbench,pmlr-v306-zhang26fi}.
These successes are supported by the extension of reinforcement learning with verifiable rewards (RLVR)~\citep{su-etal-2026-crossing,guo2025deepseek,shao2024deepseekmath}, \textit{e.g.}, by using elaborate rubrics to provide discriminative reward signals for on-policy supervision~\citep{gunjal2026rubrics,huang2025reinforcement,shen2026rethinking}.
However, as training progresses, predefined criteria may become saturated or remain unexplored (\hyperref[imgs:intro]{Figure~\ref*{imgs:intro} (a)}), resulting in stalled policy improvement~\citep{ding2026evorubrics}.

To mitigate this issue, recent work adopts a rubric-evolution paradigm that utilizes external feedback to update criteria, incorporating desired response patterns or previously overlooked errors~\citep{shao2026dr,rezaei2026online,wang2026co,ding2026evorubrics,li2026evolm}.
Despite their effectiveness, these methods overlook whether new criteria align with the task objective and the policy's capability boundary~\citep{kim2026llm}, potentially allowing superficial attributes (\textit{e.g.}, response length or format~\citep{rezaei2026online}) to drive discriminativeness.
Worse still, although an improved rubric provides more informative rewards, its outcome-only supervision forces the policy to discover desired behaviors through sample-inefficient trial and error~\mbox{\citep{wang2026textgrad,zhang2026advancing,zhou2026breaking}}.
This motivates a critical question: \textit{Can rubric's evolution track the policy's evolving capability while guiding the policy's exploration beyond scalar rewards?}

Pursuing this objective raises significant challenges.
First, \textit{how} to identify the current capability boundary?
Judging capabilities from query-only failures may confuse missing capabilities with those elicitable under rubric guidance (\hyperref[imgs:intro]{Figure~\ref*{imgs:intro} (b)}); the latter are precisely what training should internalize in the model's parameters.
Second, \textit{which} evolved criteria should be adopted?
Indiscriminately adding new criteria risks stylistic drift~\citep{yang2026rubrics}, whereas appropriate criteria should expose differences in how current responses satisfy the optimization objective.
Third, \textit{when} does rubric guidance provide a reliable direction for exploration?
Guidance may elicit different behaviors that already fall within the policy's unaided capabilities.
 
To dismantle these barriers, we propose \textbf{Co}-Evolving Policy and Rubric via \textbf{C}apability \textbf{A}wareness (\textbf{CoCA}), a closed-loop framework that couples rubric evolution with policy optimization through criterion-level capability assessment (\hyperref[imgs:intro]{Figure~\ref*{imgs:intro} (c)}).
Specifically, CoCA first sample query-only and rubric-conditioned rollouts from the same policy and contrast criterion satisfaction to characterize its current capability boundary and the associated supervision requirements.
Subsequently, CoCA selectively recalibrates criterion difficulty to expose differences in task fulfillment among distinct responses while preserving the original evaluation intent.
Finally, to provide a reliable direction for exploration, CoCA treats a conditioned response as a guided frontier only if it satisfies the diagnosed targets, using it as privileged context for token-level supervision through self-distillation.
Otherwise, optimization of conditioned rollouts expands the capability frontier, supplying improved responses for capability diagnosis and subsequent self-distillation, thereby closing the co-evolution loop.

We evaluate CoCA with Qwen3-4B/8B~\citep{yang2025qwen3}, and Gemma-3-4B-it~\citep{team2025gemma} across 7 benchmarks spanning the medical, scientific, and psychological domains. 
Extensive experiments demonstrate that CoCA achieves consistent gains over standard rubric-based RL while reaching comparable performance in approximately $2\times$ fewer training steps.
Moreover, CoCA surpasses state-of-the-art counterparts while maintaining robust out-of-domain generalization.
Further analyzes trace how behaviors elicited by rubric guidance become consolidated into query-only responses, providing evidence of capability internalization.


Our main contributions are summarized as follows:
\begin{itemize}[leftmargin=*]
    \item We propose CoCA, a novel policy-rubric co-evolutionary RL framework that persistently tracks the policy's capability boundary to maintain informative supervision throughout training.
    \item Distinct from existing isolated methods, CoCA establishes dual-group capability diagnosis as a shared decision signal for rubric adaptation and policy learning, enabling closed-loop policy--rubric co-evolution.
    \item Extensive experiments demonstrate that CoCA outperforms representative rubric-based methods in open-ended generation.
\end{itemize}

\section{Preliminaries}
\label{sec:preliminary}

\noindent\textbf{Rubric-based Reinforcement Learning}
\label{sec:preliminary_rar} extends verifiable rewards to open-ended tasks by decomposing response quality into interpretable criteria~\citep{gunjal2026rubrics,zhou2026breaking,rezaei2026online,rezaei2026rubric}.
Each query $x$ is paired with a rubric $\mathcal{R}_x=\{(c_j,w_j)\}_{j=1}^{J_x}$, where $J_x$ is the number of criteria, $c_j$ denotes the $j$-th criterion, and $w_j$ is its signed importance weight.
Given a policy response $y_i$, a fixed LLM judge $\mathcal{J}_\phi$ returns a binary verdict $z_{i,j}=\mathcal{J}_\phi(x,y_i,c_j)\in\{0,1\}$ indicating whether the criterion is met.
Positive-weight criteria reward desirable behaviors, while negative-weight criteria penalize undesirable ones.
The scalar reward aggregates these verdicts and normalizes by the total positive weight:
\begin{equation}
    r_i=\frac{\sum_{j=1}^{J_x}w_jz_{i,j}}
    {\sum_{\substack{j=1,w_j>0}}^{J_x}w_j}.
    \label{eq:rubric_reward_detailed}
\end{equation}
This normalization assumes at least one positive-weight criterion and expresses reward relative to the rubric's total positive weight.
A single positive-weight binary correctness criterion recovers the standard verifiable-reward setting.

Group Relative Policy Optimization (GRPO)~\citep{shao2024deepseekmath} uses this score as an outcome reward.
For each query, it samples $K$ responses $y_i\sim\pi_{\theta_{\mathrm{old}}}(\cdot\mid x)$, where $\theta_{\mathrm{old}}$ is the frozen pre-update snapshot of the trainable policy parameters $\theta$.
Writing $\bar r=K^{-1}\sum_{i=1}^{K}r_i$ and $\sigma_r$ for the group reward mean and standard deviation, the group-normalized advantage is $\widehat{A}_i=(r_i-\bar r)/(\sigma_r+\epsilon)$, with $\epsilon>0$ ensuring numerical stability.
The token-level importance ratio is $\rho_{i,t}(\theta)=\pi_\theta(y_{i,t}\mid x,y_{i,<t})/\pi_{\theta_{\mathrm{old}}}(y_{i,t}\mid x,y_{i,<t})$.
The clipped policy loss is
\begin{equation}
\begin{split}
    \mathcal{L}_{\mathrm{GRPO}}(\theta)
    =-\frac{1}{K}\sum_{i=1}^{K}\frac{1}{|y_i|}\sum_{t=1}^{|y_i|}
    \Big[&\min\!\left(
        \rho_{i,t}(\theta)\widehat{A}_i,
        \operatorname{clip}\!\left(\rho_{i,t}(\theta),1-\varepsilon,1+\varepsilon\right)\widehat{A}_i
    \right)\Big],
    \label{eq:rubric_grpo_loss_detailed}
\end{split}
\end{equation}
where $\varepsilon$ controls the clipping range.
Since $r_i$ is assigned at the sequence level, the same advantage $\widehat{A}_i$ supervises every token in $y_i$, without token-level attribution for individual rubric criteria.

\begin{figure*}[!th] 
    \centering 
    \begin{minipage}[t]{\linewidth} 
        \centering
        \includegraphics[width=\linewidth]{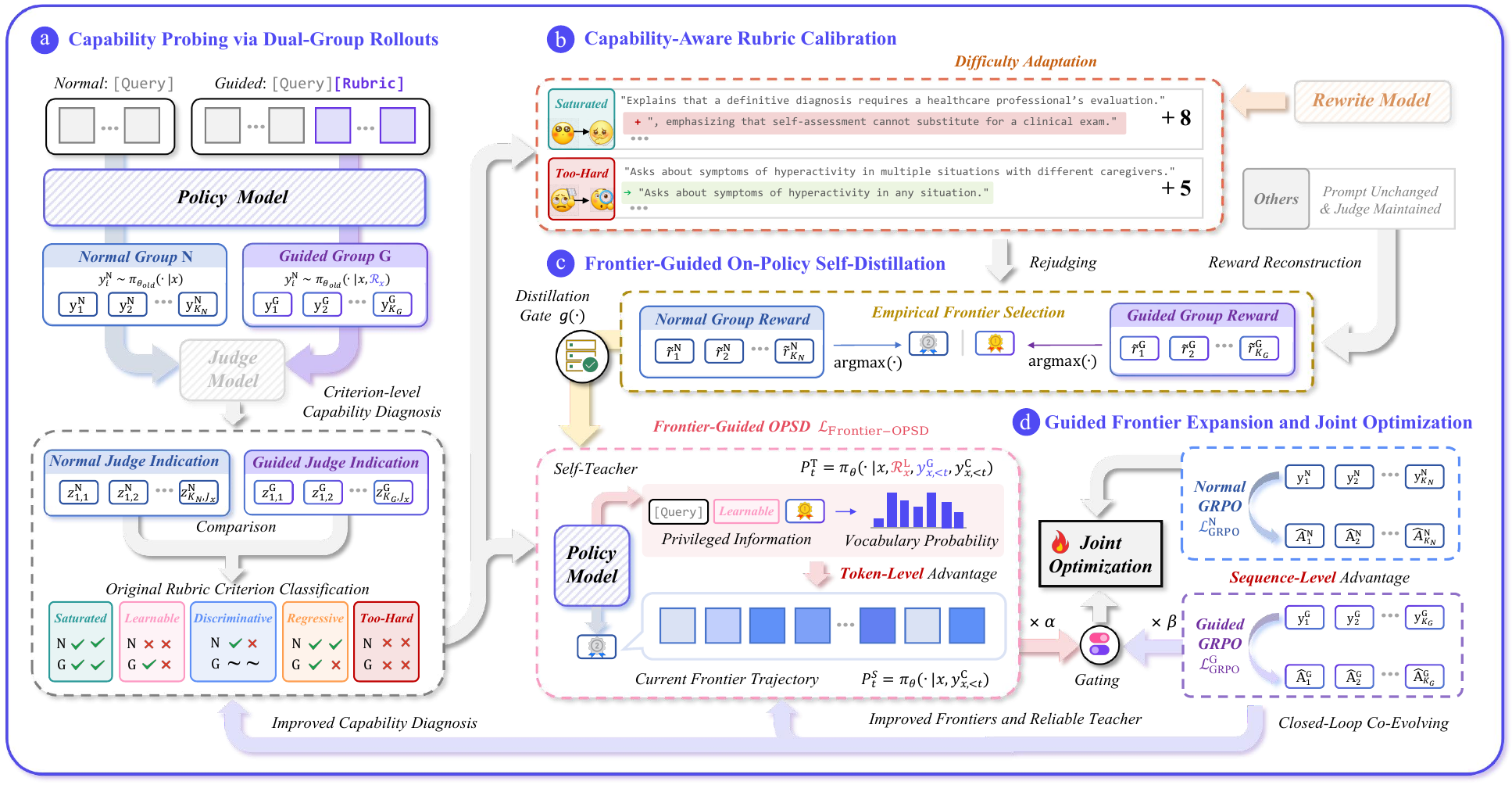} 
    \end{minipage}
   \caption{
        Pipeline overview of our CoCA.
        Dual-Group rollouts contrast query-only and rubric-conditioned behavior to diagnose criterion-level capabilities, guiding capability-aware rubric calibration and frontier selection.
        Normal GRPO remains active for every query, complemented by Frontier-Guided OPSD when a qualified frontier is available.
        Otherwise, Guided GRPO expands the rubric-conditioned frontier, supplying improved responses for the preceding two stages to close the policy--rubric co-evolution loop.
        The algorithm pseudocode of CoCA is provided in Appendix~\ref{sec:algorithm}.
    }
    \label{imgs:framework} 
      
\end{figure*}

\noindent\textbf{On-Policy Self-Distillation}
\label{sec:preliminary_opsd}
(OPSD) converts training-time privileged information into dense token-level supervision on trajectories sampled by the policy itself~\citep{zhao2026selfdistilled,li2026rethinking,han2026adaptive}.
Let $\xi$ denote information available only during training, such as a reference solution~\citep{zhao2026selfdistilled,liu2026crosslingual}, external feedback~\citep{hubotter2026reinforcement,he2026selfdistillation}, or a rubric~\citep{rezaei2026rubric,xia2026enhancing}.
For a query-only response $y\sim\pi_\theta(\cdot\mid x)$, the student and self-teacher share the same policy but receive different contexts.
At position $t$, their next-token distributions are $p_t^{\mathrm{S}}=\pi_\theta(\cdot\mid x,y_{<t})$ and $p_t^{\mathrm{T}}=\pi_\theta(\cdot\mid x,\xi,y_{<t})$, respectively.
The forward-KL objective is
\begin{equation}
    \mathcal{L}_{\mathrm{OPSD}}(\theta)
    =\mathbb{E}_{x,\,y\sim\pi_\theta(\cdot\mid x)}
    \left[
        \frac{1}{|y|}\sum_{t=1}^{|y|}
        D_{\mathrm{KL}}\!\left(
            \operatorname{sg}[p_t^{\mathrm{T}}]\,\|\,p_t^{\mathrm{S}}
        \right)
    \right],
    \label{eq:opsd_loss}
\end{equation}
where $\operatorname{sg}[\cdot]$ stops gradients through the teacher distribution.
Evaluating both distributions on student-generated prefixes aligns supervision with the policy's current state distribution.
Unlike the sequence-level GRPO signal, Eq.~\ref{eq:opsd_loss} supplies a soft target at every token, distilling behaviors elicited by $\xi$ into the query-only policy used at inference time.

\section{Methodology}
\label{sec:method}

\subsection{Capability Probing via Dual-Group Rollouts}
\label{sec:capability_probing}

\noindent\textbf{Criterion-level Capability Diagnosis.}
For each query $x$ with rubric $\mathcal{R}_x=\{(c_j,w_j)\}_{j=1}^{J_x}$, we independently sample $K_{\mathrm{N}}$ query-only responses and $K_{\mathrm{G}}$ rubric-conditioned responses from the same pre-update policy (\hyperref[imgs:framework]{Figure~\ref*{imgs:framework} (a)}), forming the \textit{Normal Group} $\mathrm{N}$ and \textit{Guided Group} $\mathrm{G}$, respectively:
\begin{equation}
\begin{aligned}
    \mathcal{Y}_x^{\mathrm{N}}&=\{y_i^{\mathrm{N}}\}_{i=1}^{K_{\mathrm{N}}},
    &y_i^{\mathrm{N}}&\sim\pi_{\theta_{\mathrm{old}}}(\cdot\mid x),\\
    \mathcal{Y}_x^{\mathrm{G}}&=\{y_i^{\mathrm{G}}\}_{i=1}^{K_{\mathrm{G}}},
    &y_i^{\mathrm{G}}&\sim\pi_{\theta_{\mathrm{old}}}(\cdot\mid x,\mathcal{R}_x).
\end{aligned}
\label{eq:dual_group_rollouts}
\end{equation}

A fixed judge $\mathcal{J}_\phi$ then evaluates each response against each criterion. Let $z_{i,j}^{v}=\mathcal{J}_\phi(x,y_i^v,c_j)\in\{0,1\}$ indicate whether $y_i^v$ meets the criterion $c_j$, where $v\in\{\mathrm{N},\mathrm{G}\}$ and $K_v$ denotes the size of group $v$.
Notably, target achievement depends on the sign of $w_j$: for a positive-weight criterion, it means satisfying the criterion, whereas for a negative-weight criterion, it means avoiding the undesirable behavior specified by the criterion.
We therefore convert judge verdicts into a polarity-invariant target-achievement indicator:
\begin{equation}
    s_{i,j}^{v}=
    \begin{cases}
        z_{i,j}^{v}, & w_j>0,\\
        1-z_{i,j}^{v}, & w_j<0,
    \end{cases}
    \qquad
    \bar{s}_j^{v}=\frac{1}{K_v}\sum_{i=1}^{K_v}s_{i,j}^{v}.
    \label{eq:target_achievement}
\end{equation}
Here, $\bar{s}_j^{v}$ measures the empirical target-achievement rate within group $v$. The achievement signature $(\bar{s}_j^{\mathrm{N}},\bar{s}_j^{\mathrm{G}})$ induces the following diagnosis at the criterion level:
{\small
\begin{equation}
    \underbrace{
        \bar{s}_j^{\mathrm{N}}=1\ \land\ \bar{s}_j^{\mathrm{G}}=1
    }_{\text{\textit{Saturated}}},
    \quad
    \underbrace{
        \bar{s}_j^{\mathrm{N}}=0\ \land\ \bar{s}_j^{\mathrm{G}}>0
    }_{\text{\textit{Learnable}}},
    \quad
    \underbrace{
        \bar{s}_j^{\mathrm{N}}=0\ \land\ \bar{s}_j^{\mathrm{G}}=0
    }_{\text{\textit{Too-Hard}}},
    \quad
    \underbrace{
        0<\bar{s}_j^{\mathrm{N}}<1
    }_{\text{\textit{Discriminative}}},
    \quad
    \underbrace{
        \bar{s}_j^{\mathrm{N}}=1\ \land\ \bar{s}_j^{\mathrm{G}}<1
    }_{\text{\textit{Regressive}}}.
    \label{eq:capability_state}
\end{equation}
}%
The five states distinguish criteria that are saturated (\textit{Saturated}), observed only under guidance (\textit{Learnable}), unmet in both sampled groups (\textit{Too-Hard}), naturally variable among \textit{Normal} responses (\textit{Discriminative}), or degraded under guidance (\textit{Regressive}). 
This policy-relative diagnosis routes each criterion to rubric calibration, frontier-based capability learning, or direct group-relative optimization.

\subsection{Capability-Aware Rubric Calibration}
\label{sec:rubric_calibration}

\noindent\textbf{Difficulty Calibration.}
Inspired by curriculum learning~\citep{bengio2009curriculum,narvekar2020curriculum}, we instantiate a criterion-level curriculum by adapting each decision boundary to the current policy rather than sequencing tasks (\hyperref[imgs:framework]{Figure~\ref*{imgs:framework} (b)}). Specifically, we \textit{strengthen} a \textit{Saturated} criterion to contract its target-achievement set, and \textit{weaken} a \textit{Too-Hard} criterion to expand this set toward a reachable intermediate target. 
The other criteria remain unchanged.
Using sampled responses, a polarity-aware rewriter revises only the criterion text while preserving its intent, polarity, weight $w_j$, and metadata. 
Denoting the resulting criterion by $c'_j$, the adapted rubric is $\mathcal{R}'_x=\{(c'_j,w_j)\}_{j=1}^{J_x}$.

\noindent
\begin{minipage}[t]{0.49\linewidth}
\begin{tcolorbox}[
    enhanced,
    equal height group=calibration-prompts,
    colback=gray!5,
    colframe=gray!40,
    boxrule=0.5pt,
    arc=1mm,
    left=1mm,right=1mm,top=1mm,bottom=1mm
]
\small
\textbf{[System]} Given $c_j^{\textit{Saturated}}$, $\mathcal{Y}_x^{\mathrm{N}}$, and $\mathcal{Y}_x^{\mathrm{G}}$, tighten the same decision boundary while preserving its intent and polarity.\\[1mm]
\textcolor{mydarkred}{\textbf{[Strengthen \textit{Saturated} Criterion]}}\\
$w_j^{\textit{Saturated}}>0$: Require more complete evidence.\\
$w_j^{\textit{Saturated}}<0$: Penalize subtler violations.
\end{tcolorbox}
\end{minipage}\hfill
\begin{minipage}[t]{0.49\linewidth}
\begin{tcolorbox}[
    enhanced,
    equal height group=calibration-prompts,
    colback=gray!5,
    colframe=gray!40,
    boxrule=0.5pt,
    arc=1mm,
    left=1mm,right=1mm,top=1mm,bottom=1mm
]
\small
\textbf{[System]} Given $c_j^{\textit{Too-Hard}}$, $\mathcal{Y}_x^{\mathrm{N}}$, and $\mathcal{Y}_x^{\mathrm{G}}$, relax the same decision boundary toward a reachable intermediate target while preserving its intent and polarity.\\[1mm]
\textcolor{darkergreen}{\textbf{[Weaken \textit{Too-Hard} Criterion]}}\\
$w_j^{\textit{Too-Hard}}>0$: Accept partial progress.\\
$w_j^{\textit{Too-Hard}}<0$: Penalize only clear violations.
\end{tcolorbox}
\end{minipage}

\noindent\textbf{Selective Rejudging and Reward Reconstruction.}
Calibration changes only a subset of criteria, so we rejudge the modified criterion--response pairs in both the \textit{Normal} and \textit{Guided Groups} while reusing all other verdicts. Let $\tilde{z}_{i,j}^{v}$ denote the verdict under $\mathcal{R}'_x$. For $v\in\{\mathrm{N},\mathrm{G}\}$,
\begin{equation}
    \tilde{z}_{i,j}^{v}=
    \begin{cases}
        \mathcal{J}_\phi(x,y_i^v,c'_j), & c'_j\neq c_j,\\
        z_{i,j}^{v}, & c'_j=c_j.
    \end{cases}
    \label{eq:selective_rejudging}
\end{equation}
Aggregating these verdicts into the reward in Eq.~\ref{eq:rubric_reward_detailed} gives the reconstructed reward for each group:
\begin{equation}
    \tilde{r}_i^{v}
    =\frac{\sum_{j=1}^{J_x}w_j\tilde{z}_{i,j}^{v}}
    {\sum_{\substack{j=1,w_j>0}}^{J_x}w_j},
    \qquad v\in\{\mathrm{N},\mathrm{G}\}.
    \label{eq:reconstructed_reward}
\end{equation}
Both groups are therefore evaluated under the same policy-adapted rubric, making their rewards directly comparable for subsequent frontier selection and policy optimization.

\subsection{Frontier-Guided On-Policy Self-Distillation}
\label{sec:frontier_opsd}

\noindent\textbf{Empirical Frontier Selection and Gating.}
We distill only capability gains that address the diagnosed gaps and exceed the policy's query-only behavior by a prescribed margin. Let
$\mathcal{I}_x^{\mathrm{L}}=\{j\in\{1,\ldots,J_x\}:
\bar{s}_j^{\mathrm{N}}=0\land\bar{s}_j^{\mathrm{G}}>0\}$
denote the indices of the \textit{Learnable} criteria, and let
$\mathcal{R}_x^{\mathrm{L}}=\{(c'_j,w_j):j\in\mathcal{I}_x^{\mathrm{L}}\}$
be the corresponding sub-rubric. We also obtain the post-calibration target-achievement indicator $\tilde{s}_{i,j}^{v}$ by replacing $z_{i,j}^{v}$ with $\tilde{z}_{i,j}^{v}$ in Eq.~\ref{eq:target_achievement}. The highest-reward \textit{Normal} response defines the \textit{Current Frontier}:
\begin{equation}
    i_x^{\mathrm{C}}=\underset{1\leq i\leq K_{\mathrm{N}}}{\arg\max}\;\tilde{r}_i^{\mathrm{N}},
    \qquad
    y_x^{\mathrm{C}}=y_{i_x^{\mathrm{C}}}^{\mathrm{N}}.
    \label{eq:current_frontier}
\end{equation}
In contrast, a \textit{Guided} response is eligible only if it achieves every \textit{Learnable} criterion. We define the set of eligible \textit{Guided} response indices $\mathcal{E}_x^{\mathrm{G}}$ as:
\begin{equation}
    \mathcal{E}_x^{\mathrm{G}}
    =
    \begin{cases}
        \left\{i\in\{1,\ldots,K_{\mathrm{G}}\}:\tilde{s}_{i,j}^{\mathrm{G}}=1
        \ \forall j\in\mathcal{I}_x^{\mathrm{L}}\right\},
        & \mathcal{I}_x^{\mathrm{L}}\neq\emptyset,\\
        \emptyset, & \text{otherwise}.
    \end{cases}
    \label{eq:guided_eligible_set}
\end{equation}
When $\mathcal{E}_x^{\mathrm{G}}\neq\emptyset$, the eligible response with the highest full-rubric reward becomes the \textit{Guided Frontier}:
\begin{equation}
    i_x^{\mathrm{G}}=\underset{i\in\mathcal{E}_x^{\mathrm{G}}}{\arg\max}\;\tilde{r}_i^{\mathrm{G}},
    \qquad
    y_x^{\mathrm{G}}=y_{i_x^{\mathrm{G}}}^{\mathrm{G}}.
    \label{eq:guided_frontier}
\end{equation}
Denoting the frontier reward gap by $\Delta_x$, we define the distillation gate as
\begin{equation}
    g(x)=\mathbbm{1}\!\left[
        \mathcal{I}_x^{\mathrm{L}}\neq\emptyset
        \ \land\ \mathcal{E}_x^{\mathrm{G}}\neq\emptyset
        \ \land\ \Delta_x>\delta
    \right],
    \qquad
    \Delta_x
    =\tilde{r}_{i_x^{\mathrm{G}}}^{\mathrm{G}}
    -\tilde{r}_{i_x^{\mathrm{C}}}^{\mathrm{N}},
    \label{eq:frontier_gate}
\end{equation}
where $\delta\geq0$ is a minimum improvement margin. This gate admits only \textit{Guided Frontiers} that satisfy all \textit{Learnable} targets and strictly improve the full-rubric reward. 

\noindent\textbf{Frontier-conditioned Token-level Distillation.}
For each active query with $g(x)=1$, the self-teacher receives the \textit{Learnable} criteria in $\mathcal{R}_x^{\mathrm{L}}$ together with the \textit{Guided Frontier} $y_x^{\mathrm{G}}$, whereas the student receives only $x$. Both share the current policy $\pi_\theta$ and are teacher-forced along the \textit{Current Frontier}:
\begin{equation}
    p_t^{\mathrm{S}}=\pi_\theta(\cdot\mid x,y_{x,<t}^{\mathrm{C}}),
    \qquad
    p_t^{\mathrm{T}}=\pi_\theta(\cdot\mid x,\mathcal{R}_x^{\mathrm{L}},
    y_x^{\mathrm{G}},y_{x,<t}^{\mathrm{C}}).
    \label{eq:frontier_teacher_student}
\end{equation}
We then minimize the forward KL divergence at each response position (\hyperref[imgs:framework]{Figure~\ref*{imgs:framework} (c)}):
\begin{equation}
    \mathcal{L}_{\mathrm{Frontier\text{-}OPSD}}(\theta;x)
    =\frac{1}{|y_x^{\mathrm{C}}|}
    \sum_{t=1}^{|y_x^{\mathrm{C}}|}
    D_{\mathrm{KL}}\!\left(
        \operatorname{sg}[p_t^{\mathrm{T}}]\,\|\,p_t^{\mathrm{S}}
    \right).
    \label{eq:frontier_opsd_loss}
\end{equation}
Crucially, $y_x^{\mathrm{G}}$ provides context rather than an imitation target: both distributions are aligned on $y_x^{\mathrm{C}}$, yielding local corrections to query-only behavior. 
For efficiency, we approximate the full-vocabulary KL with the teacher's top-$k_{\mathrm{tok}}$ tokens and one tail-mass bin, where $k_{\mathrm{tok}}$ is the retained support size.

\subsection{Guided Frontier Expansion and Joint Optimization}
\label{sec:joint_optimization}

\noindent\textbf{Group-wise Relative Optimization and Frontier Expansion.}
Although evaluated under the same adapted rubric, the two groups are generated from different contexts and can exhibit distinct reward distributions; we therefore compute relative advantages independently within each group:
\begin{equation}
    \overline{\tilde{r}}^{\,v}
    =\frac{1}{K_v}\sum_{\ell=1}^{K_v}\tilde{r}_{\ell}^{v},
    \qquad
    \widehat{A}_i^{v}
    =\frac{\tilde{r}_i^{v}-\overline{\tilde{r}}^{\,v}}
    {\sigma_{\tilde{r}}^{v}+\epsilon},
    \qquad v\in\{\mathrm{N},\mathrm{G}\},
    \label{eq:groupwise_advantages}
\end{equation}
where $\sigma_{\tilde{r}}^{v}$ is the reward standard deviation within Group $v$. Applying Eq.~\ref{eq:rubric_grpo_loss_detailed} with $K=K_v$, using query-only context $x$ for $\mathrm{N}$ and rubric-conditioned context $(x,\mathcal{R}_x)$ for $\mathrm{G}$, yields the per-query losses $\mathcal{L}_{\mathrm{GRPO}}^{\mathrm{N}}(\theta;x)$ and $\mathcal{L}_{\mathrm{GRPO}}^{\mathrm{G}}(\theta;x)$.
\textit{Normal} GRPO remains active for every query, whereas \textit{Guided} GRPO is masked by $1-g(x)$, so all $K_{\mathrm{G}}$ Guided responses contribute only when $g(x)=0$.
This restriction directs \textit{Guided} GRPO toward queries without a qualified frontier.

\noindent\textbf{Joint Training Objective.}
As shown in \hyperref[imgs:framework]{Figure~\ref*{imgs:framework} (d)}, with coefficients $\alpha$ and $\beta$ for Frontier-OPSD and Guided GRPO, respectively, the joint objective is defined as:
\begin{equation}
    \mathcal{L}_{\mathrm{joint}}(\theta)=\mathbb{E}_{x}\!\left[\mathcal{L}_{\mathrm{GRPO}}^{\mathrm{N}}(\theta;x)+\alpha g(x)\mathcal{L}_{\mathrm{Frontier\text{-}OPSD}}(\theta;x)+\beta\bigl(1-g(x)\bigr)\mathcal{L}_{\mathrm{GRPO}}^{\mathrm{G}}(\theta;x)\right].
    \label{eq:joint_objective}
\end{equation}
Here, $\mathcal{L}_{\mathrm{GRPO}}^{\mathrm{N}}$ improves query-only behavior, while the gated branches use $\mathcal{L}_{\mathrm{GRPO}}^{\mathrm{G}}$ to build a qualified frontier when $g(x)=0$ and $\mathcal{L}_{\mathrm{Frontier\text{-}OPSD}}$ to transfer it when $g(x)=1$. \textit{Guided} GRPO then yields stronger rubric-conditioned rollouts, which refresh capability diagnosis, recalibrate the rubric, and supply improved frontiers to subsequent OPSD, closing the policy--rubric co-evolution loop.

\begin{table*}[!th]
    \vspace{5pt}
    \centering
    \caption{ 
    Main results across seven benchmarks spanning the medical, science, and psychological domains.
    The best scores are highlighted in  \textbf{bold}, while the second-best scores are \underline{underlined}. 
    All scores are normalized to $0 \sim 100$.
    Additional experiments extending CoCA to larger model and other GRPO variants are presented in \hyperref[sec:larger_models]{Appendix~\ref*{sec:larger_models}}  and \hyperref[sec:grpo_variants]{Appendix~\ref*{sec:grpo_variants}}, respectively.
    } 
    \vspace{-5pt}
    \label{tab:unibench}
    \resizebox{1\linewidth}{!}{
    \begin{tabular}{l ccc cc cc c}
        \whline
        \multirow{2}{*}{\textbf{Method}} 
        &\multicolumn{3}{c}{\textbf{Medical Domain}} 
        & \multicolumn{2}{c}{\textbf{Science Domain}} 
        & \multicolumn{2}{c}{\textbf{Psychological Domain}} 
        
        & \multirow{2}{*}{\textbf{Mean}} \\
        
        \cmidrule(r){2-4}  \cmidrule(lr){5-6} 
        \cmidrule(l){7-8}  
        
        ~ &
        HealthBench &  RaR-Medicine &  LLMEval-Med & 
        ResearchQA  &  RaR-Science  & 
        CPsyCoun  &  HeartBench
        ~ \\
        \hline

        \rowcolor{mygray} Qwen3-4B 
        & $22.38$ & $51.78$ & $38.38$ & $59.93$ & $69.76$ & $31.39$ & $44.17$ &  $45.40$  \\
        LLM-Rank  	
        & $30.10$ & $52.54$ & $42.13$ & $73.09$ & $\underline{74.91}$ & $33.56$ & $43.54$ &  $49.98$ \\
        RaR
        & $37.12$ & $56.03$ & $42.58$ & $78.42$ & $73.81$ & $77.41$  & $\mathbf{50.14}$ &  $59.36$  \\
        RuscaRL 	
        & $\underline{40.44}$ & $\mathbf{58.37}$ & $43.63$ & $\underline{79.57}$ & $73.12$ & $\underline{82.03}$ & $45.74$ & $\underline{60.41}$ \\
        Online Rubrics
        & $37.77$ & $56.28$ & $\underline{44.68}$ & $78.96$ & $74.35$ & 	$79.55$  & $44.92$ & $59.50$ \\
        RGSD  	
        & $18.67$ & $49.29$ & $37.48$ & $61.53$ & $67.44$ & $27.35$&  $44.03$ & $43.68$ \\
        \rowcolor{mypurple}
        CoCA (Ours)
        & $\mathbf{41.01}$ & $\underline{56.33}$ & $\mathbf{46.48}$ & $\mathbf{79.96}$ & $\mathbf{76.56}$ & $\mathbf{84.48}$ & $\underline{48.27}$  & $\mathbf{61.87}$  \\
        	
        \hline

        \rowcolor{mygray} Qwen3-8B  	
        & $25.86$ & $59.44$ & $44.28$ & $64.44$ & $73.34$ & $40.12$  & $46.29$ &  $50.54$  \\
        LLM-Rank  	
        & $36.03$ & $61.24$ & $49.63$ & $75.85$ & $75.39$ & $43.90$ & $46.51$ & $55.51$  \\
        RaR
        & $41.88$ & $64.29$ & $\underline{49.93}$ & $81.14$ & $\underline{78.63}$ & $83.71$ & $48.25$ & $63.98 $\\
        RuscaRL
        & $\underline{46.14}$ & $64.07$ & $45.28$ & $80.88$ & $77.78$ & $\underline{86.37}$ &  $51.93$ & $64.64$ \\
        Online Rubrics
        & $45.32$ & $\underline{64.59}$ & $49.48$ & $\underline{81.45}$ & $78.38$ &  $85.94$ & $\underline{52.40}$ & $\underline{65.37}$ \\
        RGSD
        & $21.38$ & $56.86$ & $39.28$ & $66.03$ & $72.13$ & $43.36$ & 	$47.37$ & $49.49$ \\
        \rowcolor{mypurple}
        CoCA (Ours)
        & $\mathbf{46.86}$ & $\mathbf{65.37}$ & $\mathbf{52.77}$ &  $\mathbf{81.55}$ &  $\mathbf{79.16}$   & $\mathbf{86.49}$  &  $\mathbf{52.67}$ & $\mathbf{66.41}$  \\
        \hline

        \rowcolor{mygray} Gemma-3-4B-it
        & $28.55$ & $45.40$ & $22.79$ & $62.49$ & $49.64$ & $30.23$ & $45.60$ &  $40.67$ \\
        LLM-Rank
        &$37.12$& $45.13$& $\underline{26.09}$ & $65.67$ & $52.79$ & $52.33$ & $48.29$ & $46.77$ \\
        RaR
        &  $48.26$ & $48.69$ & $23.69$ & $74.33$ & $52.86$ & $73.55$ & $\mathbf{49.47}$ & $52.98$ \\
        RuscaRL
        & $44.09$ & $46.89$ & $25.49$ & $75.40$ & $47.24$ & $75.07$ & $34.01$ & $49.74$ \\
        Online Rubrics
        & $\underline{50.39}$ & $\underline{49.90}$ & $23.84$ & $\underline{75.48}$ & $\mathbf{53.95}$ &  $\underline{79.10}$  & $47.59$ & $\underline{54.32}$ \\
        RGSD
        & $28.37$ & $42.73$ & $23.09$ & $65.78$ & $48.48$ & $41.26$ & $44.56$ & $42.04$  \\
        \rowcolor{mypurple}
        CoCA (Ours)
        & $\mathbf{52.71}$ &  $\mathbf{52.01}$  & $\mathbf{26.54}$  &$ \mathbf{75.97}$ & $\underline{53.13}$  & $\mathbf{83.39}$ & $\underline{48.81}$ &  $\mathbf{56.08}$ \\
        \hline

        \whline
        \end{tabular}
    }
\end{table*}

\section{Experiments}
\label{sec:experiments}

\subsection{Experimental Setup}
\label{sec:exp_setting}

\noindent \textbf{Model Architecture and Datasets.}  
We evaluate our CoCA on Qwen3-4B, Qwen3-8B~\citep{yang2025qwen3}, and Gemma-3-4B-it~\citep{team2025gemma}. We consider training and evaluation across three domains: medical, science, and psychology. 
For medical, we draw $4,000$ training examples from \textbf{HealthBench}~\citep{arora2025healthbench}, excluding HealthBench-Hard, and evaluate on HealthBench-Hard, \textbf{RaR-Medicine}~\citep{gunjal2026rubrics}, and \textbf{LLMEval-Med}~\citep{zhang-etal-2025-llmeval}. 
For science, we randomly sample $4,000$ \textbf{ResearchQA}~\citep{yifei2026researchqa} examples for training and evaluate on its validation split and the \textbf{RaR-Science} test set~\citep{gunjal2026rubrics}. 
For psychology, we construct LLM-generated rubrics for \textbf{CPsyCoun}~\citep{zhang2024cpsycoun}, randomly select $2,400$ examples for training, reserve 500 held-out examples for in-domain testing, and evaluate cross-dataset generalization on \textbf{HeartBench}~\citep{liu2025heartbench}. 
We also assess Out-of-Domain (OOD) instruction-following generalization on \textbf{InfoBench}~\citep{qin2024infobench}, \textbf{FollowBench}~\citep{jiang2024followbench}, and \textbf{AdvancedIF}~\citep{he2026advancedif}. 
Detailed descriptions and metrics for each dataset are provided in \hyperref[sec:datasets_details]{Appendix~\ref*{sec:datasets_details}}.

\noindent \textbf{Baseline.}  
We compare our CoCA against five baselines covering alternative reward construction and distinct intervention points in rubric-based RL. \textbf{LLM-Rank}~\citep{liusie-etal-2024-llm} is a rubric-free GRPO baseline that assigns relative ranks within each group, thereby ensuring reward variation. 
\textbf{RaR}~\citep{gunjal2026rubrics} serves as the standard rubric-based GRPO baseline. \textbf{RuscaRL}~\citep{zhou2026breaking}, \textbf{Online Rubrics}~\citep{rezaei2026online}, and \textbf{RGSD}~\citep{rezaei2026rubric} represent guided rollout exploration, online rubric evolution, and self-distillation, respectively.

\begin{table*}[!th]
    \vspace{-5pt}
    \centering
    \small
    \setlength{\tabcolsep}{5pt}
    \caption{Experiment results of rubric-conditioned in-domain capability and out-of-domain instruction-following generalization. All compared models are trained on HealthBench. $^\dagger$: We provide the full ground-truth rubric together with each HealthBench-Hard query.
    }
      \vspace{-5pt}
    \label{tab:ood_results}
    \begin{tabular}{l ccc ccc ccc}
        \whline
        
        \multirow{2}{*}{\textbf{Dataset}}
        & \multicolumn{3}{c}{\textbf{Qwen3-4B}} & \multicolumn{3}{c}{\textbf{Qwen3-8B}} & \multicolumn{3}{c}{\textbf{Gemma-3-4B-it}}\\

        \cmidrule(r){2-4} \cmidrule(lr){5-7} \cmidrule(l){8-10}
  
         ~ &
         Base & RuscaRL & CoCA &
         Base & RuscaRL & CoCA &
         Base & RuscaRL & CoCA \\

        \hline
\multirow{2}{*}{HealthBench $^{\dagger}$}
& \multirow{2}{*}{$62.96$} & $75.54$ & $\mathbf{80.34}$
& \multirow{2}{*}{$71.83$} & $81.13$ & $\mathbf{83.02}$
& \multirow{2}{*}{$65.50$} & $74.94$ & $\mathbf{77.79}$ \\
& & {\scriptsize\textcolor{mygreen}{$+12.58$}} & {\scriptsize\textcolor{mygreen}{$+17.38$}}
& & {\scriptsize\textcolor{mygreen}{$+9.30$}} & {\scriptsize\textcolor{mygreen}{$+11.19$}}
& & {\scriptsize\textcolor{mygreen}{$+9.44$}} & {\scriptsize\textcolor{mygreen}{$+12.29$}} \\
        \hline

\multirow{2}{*}{InfoBench}
& \multirow{2}{*}{$77.68$} & $75.68$ & $\mathbf{79.02}$
& \multirow{2}{*}{$78.66$} & $78.04$ & $\mathbf{80.40}$
& \multirow{2}{*}{$76.66$} & $76.44$ & $\mathbf{77.11}$ \\
& & {\scriptsize\textcolor{mydarkred}{$-2.00$}} & {\scriptsize\textcolor{mygreen}{$+1.34$}}
& & {\scriptsize\textcolor{mydarkred}{$-0.62$}} & {\scriptsize\textcolor{mygreen}{$+1.74$}}
& & {\scriptsize\textcolor{mydarkred}{$-0.22$}} & {\scriptsize\textcolor{mygreen}{$+0.45$}} \\

\multirow{2}{*}{FollowBench}
& \multirow{2}{*}{$47.92$} & $45.73$ & $\mathbf{51.58}$
& \multirow{2}{*}{$52.19$} & $48.04$ & $\mathbf{52.56}$
& \multirow{2}{*}{$59.51$} & $58.65$ & $\mathbf{60.73}$ \\
& & {\scriptsize\textcolor{mydarkred}{$-2.19$}} & {\scriptsize\textcolor{mygreen}{$+3.66$}}
& & {\scriptsize\textcolor{mydarkred}{$-4.15$}} & {\scriptsize\textcolor{mygreen}{$+0.37$}}
& & {\scriptsize\textcolor{mydarkred}{$-0.86$}} & {\scriptsize\textcolor{mygreen}{$+1.22$}} \\

\multirow{2}{*}{AdvancedIF}
& \multirow{2}{*}{$26.58$} & $25.07$ & $\mathbf{26.93}$
& \multirow{2}{*}{$32.42$} & $28.58$ & $\mathbf{33.56}$
& \multirow{2}{*}{$31.32$} & $30.79$ & $\mathbf{33.59}$ \\
& & {\scriptsize\textcolor{mydarkred}{$-1.51$}} & {\scriptsize\textcolor{mygreen}{$+0.35$}}
& & {\scriptsize\textcolor{mydarkred}{$-3.84$}} & {\scriptsize\textcolor{mygreen}{$+1.14$}}
& & {\scriptsize\textcolor{mydarkred}{$-0.53$}} & {\scriptsize\textcolor{mygreen}{$+2.27$}} \\

        \whline
    \end{tabular}
      \vspace{-5pt}
\end{table*}

\noindent \textbf{Implementation Details.}  
Unless otherwise specified, we use equal rollout group sizes $K_{\mathrm{N}}=K_{\mathrm{G}}=4$ for the \textit{Normal} and \textit{Guided Groups} (budget expansion study is presented in \hyperref[sec:budget]{Appendix~\ref*{sec:budget}}), 
set $\alpha=0.001$ and $\beta=0.1$. 
DeepSeek-V3.2~\citep{liu2025deepseek} serves as the fixed criterion-level judge and rewrite model. 
Full implementation and prompt details are provided in \hyperref[sec:implementation_details]{Appendix~\ref*{sec:implementation_details}} and \hyperref[sec:implementation_details]{Appendix~\ref*{sec:prompt_design}}, respectively.

\subsection{Overall Performance}
\label{sec:over_results}

\noindent \textbf{CoCA demonstrates robust effectiveness and generalizability across backbones and domains.}
As shown in Table~\ref{tab:unibench}, CoCA achieves the highest mean score for Qwen3-4B, Qwen3-8B, and Gemma-3-4B-it ($61.87\%$, $66.41\%$, and $56.08\%$), improving over standard rubric-based GRPO (RaR) by \textcolor{mygreen}{$+2.51\%$}, \textcolor{mygreen}{$+2.43\%$}, and \textcolor{mygreen}{$+3.10\%$}, respectively.
It also outperforms RuscaRL, Online Rubrics, and RGSD on most individual benchmarks.
Together, these results indicate that coordinating capability diagnosis, rubric calibration, guided frontier expansion, and selective transfer is more effective than repairing individual stages of rubric-based RL in isolation.
The weaker LLM-Rank results suggest that holistic group rankings discard useful criterion-level structure, whereas RGSD's weak results are consistent with recent findings that unfiltered or standalone OPSD can become unstable under noisy teacher signals and teacher--student competence mismatch~\citep{xia2026enhancing,kim2026does}.

\noindent \textbf{CoCA couples stronger rubric-conditioned capability with robust OOD generalization.}
To examine whether rubric-guided gains transfer to query-only instruction following beyond the training domain, we evaluate both rubric-conditioned capability and OOD generalization.
As shown in \hyperref[tab:ood_results]{Table~\ref*{tab:ood_results}}, when the full ground-truth rubric is included in each HealthBench-Hard prompt, CoCA substantially improves over the base model across all backbones (\textit{e.g.}, \textcolor{mygreen}{$+17.38\%$} for Qwen3-4B).
It also consistently surpasses RuscaRL by \textcolor{mygreen}{$+1.89\%\sim+4.80\%$}
Across the OOD benchmarks, CoCA consistently outperforms RuscaRL and generally improves over the corresponding base models, whereas RuscaRL degrades OOD performance with drops ranging from \textcolor{mydarkred}{$-0.22\%$} to \textcolor{mydarkred}{$-4.15\%$}.
This contrast demonstrates that CoCA enables broadly transferable rubric-guided learning.

\subsection{Ablation Study}
\label{sec:ablation}


\noindent\leavevmode%

\begin{wraptable}{R}{0.55\linewidth}
    \centering
    \caption{Ablation of evaluation judge models for Qwen3-8B in the medical domain.}
    \vspace{-6pt}
    \label{tab:ablation_judge}
    \small
    \setlength{\tabcolsep}{2pt}
    \resizebox{\linewidth}{!}{%
    \begin{tabular}{llcccc}
        \toprule
        \textbf{Eval Judge}
        & \textbf{Method}
        & \textbf{HealthBench}
        & \textbf{RaR-Med.}
        & \textbf{LLMEval.}
        & \textbf{Mean} \\
        \midrule

        \multirow{3}{*}{\shortstack[l]{Qwen3.7-Flash \\ \citep{qwen37}}}
             & \cellcolor{mygray}Base & \cellcolor{mygray}$27.31$ & \cellcolor{mygray}$54.56$ & \cellcolor{mygray}$59.22$ & \cellcolor{mygray}$47.03$ \\
             & RaR & $43.04$ & $60.06$ & $62.52$ & $55.21$ \\
             & \cellcolor{mypurple}CoCA & \cellcolor{mypurple}$\mathbf{47.89}$ & \cellcolor{mypurple}$\mathbf{61.35}$ & \cellcolor{mypurple}$\mathbf{65.53}$ & \cellcolor{mypurple}$\mathbf{58.26}$ \\
        \midrule
        \multirow{3}{*}{\shortstack[l]{Kimi-K2.5 \\ \citep{team2026kimi}}}
             & \cellcolor{mygray}Base & \cellcolor{mygray}$17.93$ & \cellcolor{mygray}$55.29$ & \cellcolor{mygray}$25.04$ & \cellcolor{mygray}$32.75$ \\
             & RaR & $36.77$ & $61.79$ & $28.04$ & $42.20$ \\
             & \cellcolor{mypurple}CoCA & \cellcolor{mypurple}$\mathbf{41.59}$ & \cellcolor{mypurple}$\mathbf{63.02}$ & \cellcolor{mypurple}$\mathbf{30.13}$ & \cellcolor{mypurple}$\mathbf{44.91}$ \\
        \midrule
        \multirow{3}{*}{\shortstack[l]{GLM-5 \\ \citep{zeng2026glm}}}
             & \cellcolor{mygray}Base & \cellcolor{mygray}$12.38$ & \cellcolor{mygray}$58.74$ & \cellcolor{mygray}$36.88$ & \cellcolor{mygray}$36.00$ \\
             & RaR & $35.41$ & $60.36$ & $41.38$ & $45.72$ \\
             & \cellcolor{mypurple}CoCA & \cellcolor{mypurple}$\mathbf{40.89}$ & \cellcolor{mypurple}$\mathbf{64.35}$ & \cellcolor{mypurple}$\mathbf{43.93}$ & \cellcolor{mypurple}$\mathbf{49.72}$ \\
        \bottomrule
    \end{tabular}%
    }
    \parbox{\linewidth}{\footnotesize\raggedright \textit{Note:} ``RaR-Med.'' and ``LLMEval.'' denote the RaR-Medicine and LLMEval-Med benchmark, respectively.}
    
\end{wraptable}
\textbf{Robustness to Evaluation Judges.}
To assess the robustness of our CoCA to evaluator choice and potential reward hacking~\citep{gao2023scaling,skalse2022defining}, we evaluate Qwen3-8B in the medical domain using three alternative judges while retaining DeepSeek-V3.2 for training.
As shown in \hyperref[tab:ablation_judge]{Table~\ref*{tab:ablation_judge}}, CoCA improves the mean score over the base model by \textcolor{mygreen}{$+11.23\%$}, \textcolor{mygreen}{$+12.16\%$}, and \textcolor{mygreen}{$+13.72\%$} under the three evaluation judges, respectively. 
It consistently surpasses RaR, with absolute gains of \textcolor{mygreen}{$+2.71\%\sim+4.00\%$} in mean score.
This consistent advantage under evaluators uninvolved in training supports the transferability of CoCA's improvements across scoring models, suggesting that its gains are not solely attributable to adaptation to the training judge's preferences.


\begin{wraptable}{R}{0.45\linewidth}
    \centering
    \caption{Ablation study of the CoCA's components on Qwen3-8B.
    }
    \label{tab:ablation_component}
    \vspace{-5pt}
    \setlength{\tabcolsep}{2pt}
    \resizebox{\linewidth}{!}{%
    \begin{tabular}{ccccc}
        \whline
        
        \textbf{CARC}
        & $\mathbf{\mathcal{L}_{\mathrm{Frontier\text{-}OPSD}}}$
        & $\mathbf{\mathcal{L}_{\mathrm{GRPO}}^{\mathrm{G}}}$
        & \textbf{HealthBench}
        & \textbf{LLMEval.} \\
        
        \hline
        
        \rowcolor{mygray}
        $\times$ & $\times$ & $\times$ 
        & $41.88$ & $49.93$ \\

        $\checkmark$ & $\times$ & $\times$ 
        & $46.30$ & $48.73$ \\

        $\times$ & $\checkmark$ & $\times$ 
        & $42.39$ & $50.52$ \\

        $\checkmark$ & $\checkmark$ & $\times$
        & $45.01$ & $51.12$ \\

        $\checkmark$ & $\times$ & $\checkmark$ 
        & $44.47$ & $52.62$ \\

        $\times$ & $\checkmark$ & $\checkmark$
        & $43.11$ & $50.97$ \\

        \rowcolor{mypurple}
        $\checkmark$ & $\checkmark$ & $\checkmark$  
        & $\mathbf{46.86}$ & $\mathbf{52.77}$ \\

        \whline
    \end{tabular}%
    }
      \vspace{-6pt}
\end{wraptable}

\noindent \textbf{Component Complementarity.}
We next ablate capability-aware rubric calibration (CARC, see Section~\ref{sec:rubric_calibration}), $\mathcal{L}_{\mathrm{Frontier\text{-}OPSD}}$, and $\mathcal{L}_{\mathrm{GRPO}}^{\mathrm{G}}$ to examine how the three components interact.
As shown in \hyperref[tab:ablation_component]{Table~\ref*{tab:ablation_component}}, the full CoCA reaches $46.86\%$ on HealthBench-Hard and $52.77\%$ on LLMEval-Med, outperforming RaR by \textcolor{mygreen}{$+4.98\%$} and \textcolor{mygreen}{$+2.84\%$}, respectively.
Ablating Rubric Calibration, $\mathcal{L}_{\mathrm{Frontier\text{-}OPSD}}$, and $\mathcal{L}_{\mathrm{GRPO}}^{\mathrm{G}}$ changes HealthBench-Hard performance by \textcolor{mydarkred}{$-3.75\%$}, \textcolor{mydarkred}{$-2.39\%$}, and \textcolor{mydarkred}{$-1.85\%$}, and LLMEval-Med performance by \textcolor{mydarkred}{$-1.80\%$}, \textcolor{mydarkred}{$-0.15\%$}, and \textcolor{mydarkred}{$-1.65\%$}.
These task-dependent effects suggest complementary rather than uniformly additive roles: Rubric Calibration maintains informative objectives, $\mathcal{L}_{\mathrm{GRPO}}^{\mathrm{G}}$ expands the guided frontier, and $\mathcal{L}_{\mathrm{Frontier\text{-}OPSD}}$ selectively transfers qualified gains.
Their coordination yields the strongest overall configuration, supporting the closed-loop design of CoCA.
Further ablations are provided in \hyperref[sec:additional_experiments]{Appendix~\ref*{sec:additional_experiments}}.

\begin{figure*}[!th] 
    \centering 
    \begin{minipage}[t]{\linewidth} 
        \centering
        \includegraphics[width=\linewidth]{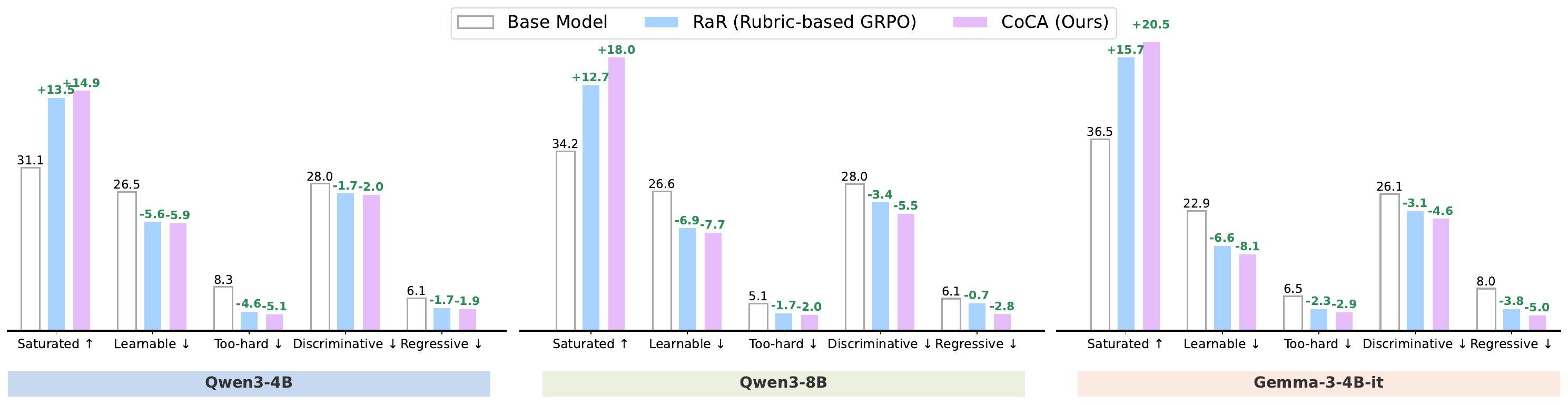} 
    \end{minipage}

    \caption{
        Overall distribution of rubric capability states on HealthBench-Hard. For each query, we sample two groups of responses: one under the query-only condition and the other with the full rubric additionally provided. Each rubric is then assigned to one of five capability states according to its satisfaction patterns across the two groups.
    }
    \label{fig:rubric_classify}
    
\end{figure*}

\begin{table*}[!th]
    \vspace{-5pt}
    \centering
   \caption{Rubric capability-state transitions on HealthBench-Hard. For each backbone, we record each rubric's \textit{Source State} under the Base model and its \textit{Target State} under a HealthBench-trained model, and report the types and proportions (\%) of the resulting transitions.
    }
      \vspace{-5pt}
    \label{tab:analysis_trans}
    \resizebox{1\linewidth}{!}{
    \begin{tabular}{c l cc cc cc}
        \whline

        \multirow{2}{*}{\textbf{Type}} &
        \multirow{2}{*}{\shortstack{\textbf{Transition} \\ \textbf{(Source $\Longrightarrow$ Target)}}}
        & \multicolumn{2}{c}{\textbf{Qwen3-4B}} & \multicolumn{2}{c}{\textbf{Qwen3-8B}} & \multicolumn{2}{c}{\textbf{Gemma-3-4B-it}}\\

        \cmidrule(r){3-4} \cmidrule(lr){5-6} \cmidrule(l){7-8} 
  
         ~ & ~ &  
         RaR  &     Ours & 
         RaR  &      Ours &
         RaR  &     Ours  \\

        \hline
        
        \multirow{4}{*}{Internalization $\uparrow$}  
        
& Learnable ${\color{mygreen}\Longrightarrow}$ Saturated 
& $8.10$ & $\mathbf{9.83}$ &  $6.96$ & $\mathbf{14.02}$ & $13.84$ &  $\mathbf{15.76}$ \\

~ & Learnable ${\color{mygreen}\Longrightarrow}$ Discriminative 
& $33.69$ & $\mathbf{34.47}$ & $32.56$ & $\mathbf{36.88}$ & $30.97$ & $\mathbf{35.70}$  \\

~ & Too-hard ${\color{mygreen}\Longrightarrow}$ Saturated 
& $2.63$ & $\mathbf{3.65}$ & $0.49$ & $\mathbf{5.25}$ & $6.34$ &  $\mathbf{8.67}$    \\

~ & Discriminative ${\color{mygreen}\Longrightarrow}$ Saturated 
& $41.17$ & $\mathbf{43.61}$ & $40.68$ &  $\mathbf{51.21}$  & $42.89$ &  $\mathbf{45.42}$  \\

        \hline

        \multirow{2}{*}{Degradation $\downarrow$}  
        
& Saturated ${\color{mydarkred}\Longrightarrow}$ Others
& $12.86$ & $\mathbf{12.21}$  & $\mathbf{8.95}$ &  $12.48$  & $13.33$  &  $\mathbf{12.50}$   \\

~ & Learnable ${\color{mydarkred}\Longrightarrow}$ Too-hard 
& $2.74$ & $\mathbf{2.23}$  & $2.26$ & $\mathbf{2.19}$ & $5.02$  & $\mathbf{3.91}$  \\

        \whline
        \end{tabular}
    }
\end{table*}

\begin{figure*}[h]
    \centering
    \begin{minipage}[t]{\textwidth}
        \centering
        \begin{subfigure}[t]{0.32\linewidth}
            \centering
            \includegraphics[width=\linewidth]{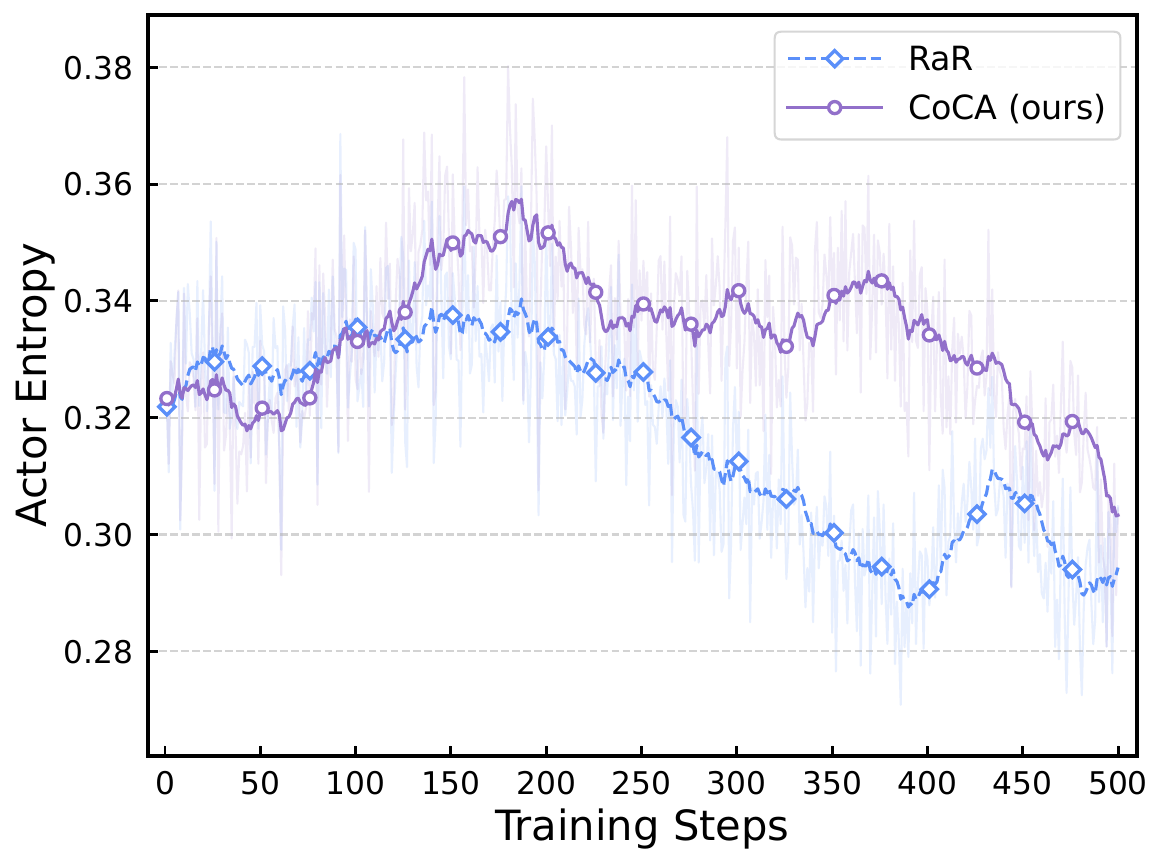}
            \caption{Training Policy Entropy}
            \label{fig:combined_actor_entropy}
        \end{subfigure}%
        \hfill
        \begin{subfigure}[t]{0.32\linewidth}
            \centering
            \includegraphics[width=\linewidth]{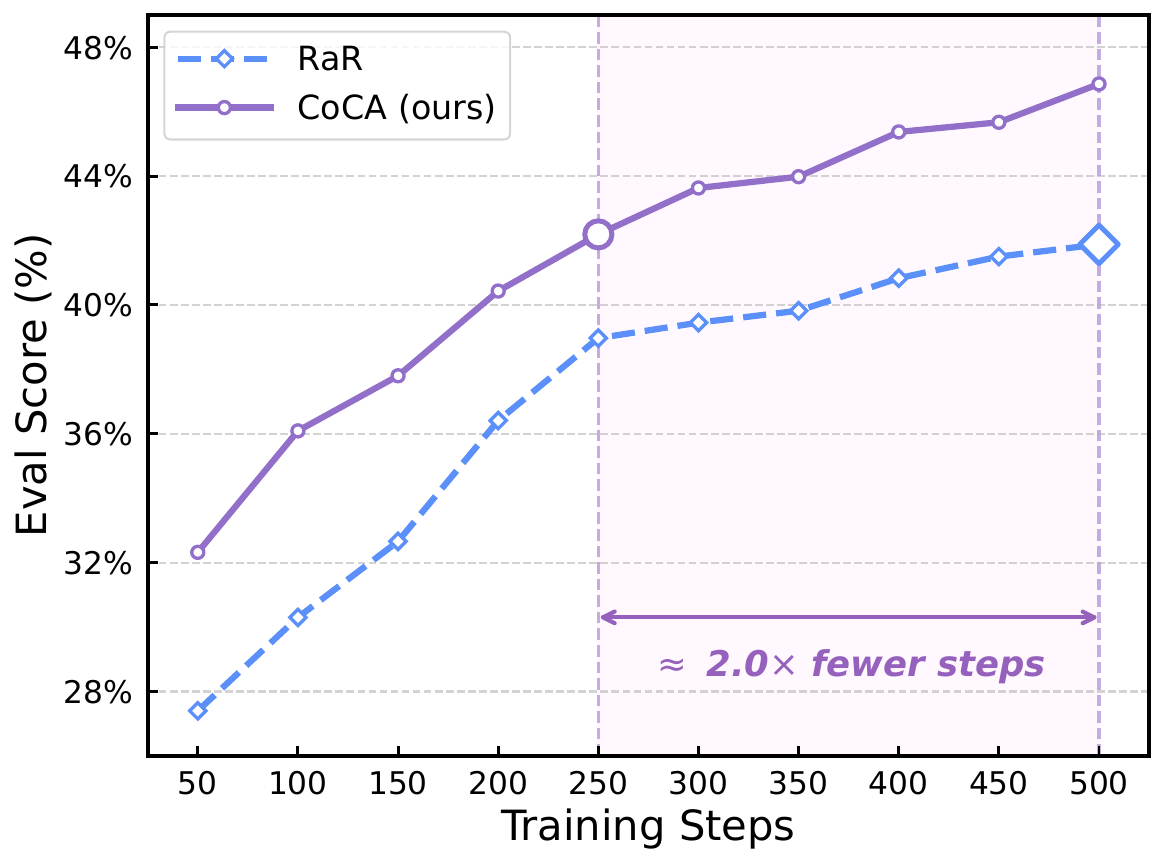}
            \caption{Training Progress}
            \label{fig:combined_sample_efficiency}
        \end{subfigure}
        \hfill
        \begin{subfigure}[t]{0.32\textwidth}
            \centering
            \includegraphics[width=\linewidth]{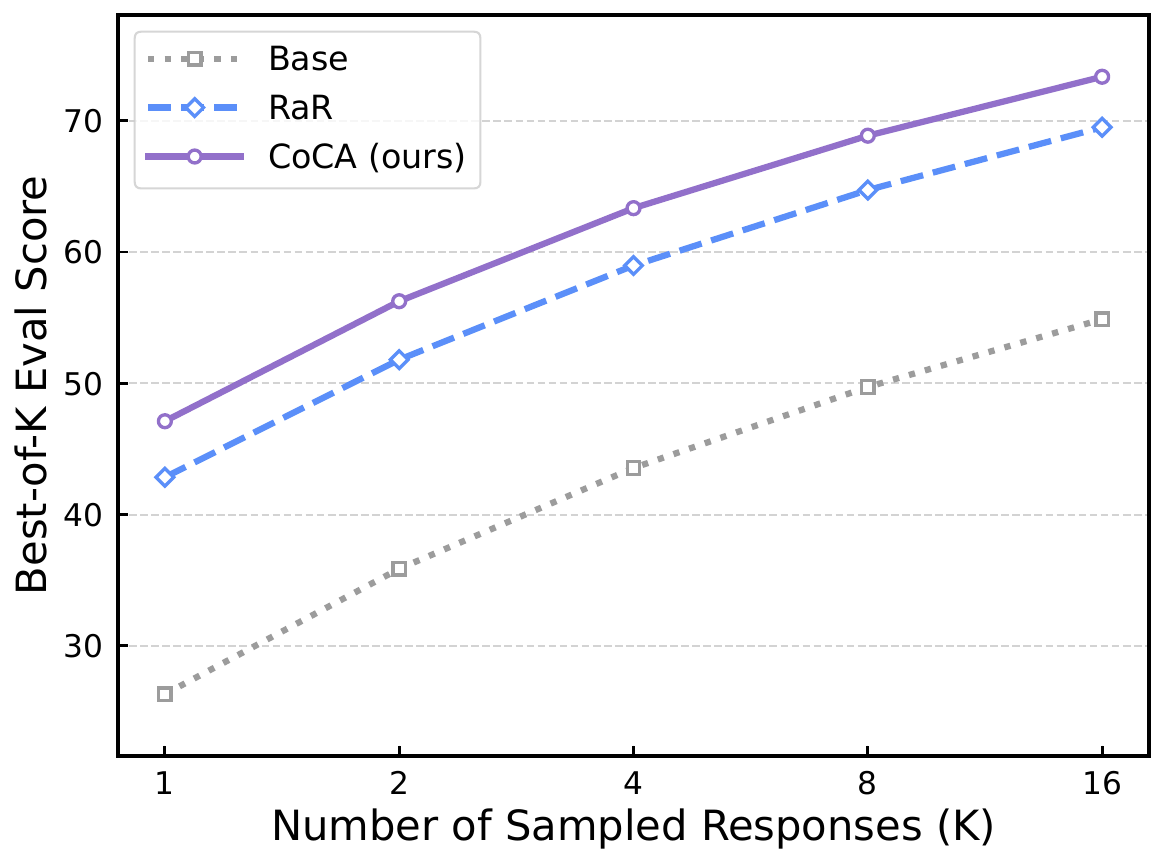}
            \caption{Best-of-K Performance}
            \label{fig:combined_best_of_k}
        \end{subfigure}
    \end{minipage}%
        
    \caption{Training dynamics and inference-time scaling of CoCA on HealthBench. 
    (a) CoCA maintains higher entropy, preserving exploration during training;
    (b) CoCA matches RaR's best performance with fewer training steps and time (\hyperref[sec:training_time_overhead]{Appendix~\ref*{sec:training_time_overhead}}); 
    (c) its consistent advantage over the base model and RaR across $K$ indicates a higher capability ceiling accessible through sampling.}

    \label{fig:training_dynamic_and_best_of_k}
\end{figure*}

\section{Analysis and Discussions}
\label{sec:analysis}

In this section, we mainly investigate: (1) \textit{What} does the distribution of rubric capability states reveal about the policy's evolving capability boundary;  (2) \textit{Which} capability-state transitions toward internalization are more pronounced under CoCA than RaR; and (3) \textit{How} does CoCA affect training dynamics and inference-time scaling?
We further discuss rubric discrimination in \hyperref[sec:rubric_discrimination]{Appendix~\ref*{sec:rubric_discrimination}}.

\noindent \textbf{Capability-state distributions reveal an evolving policy frontier.}
\hyperref[fig:rubric_classify]{Figure~\ref*{fig:rubric_classify}} shows that, relative to the base models, CoCA shifts the proportions of \textit{Saturated} and \textit{Learnable} by \textcolor{mygreen}{$+14.9\%\sim+20.5\%$} and \textcolor{mygreen}{$-8.1\%\sim-5.9\%$}, respectively, across all backbones.
The \textit{Too-Hard} and \textit{Regressive} proportions also decrease consistently, indicating fewer unreachable targets and guidance-induced failures. 
This separation is most pronounced on Qwen3-8B, where CoCA changes \textit{Saturated} and \textit{Regressive} by \textcolor{mygreen}{$+18.00\%$} and \textcolor{mygreen}{$-2.80\%$}, compared with \textcolor{mygreen}{$+12.70\%$} and \textcolor{mygreen}{$-0.70\%$} for RaR.
This redistribution suggests a narrowing gap between elicited and default capability, but marginal state counts alone cannot distinguish internalization from degradation, motivating a transition-level analysis.

\noindent \textbf{CoCA promotes capability internalization across backbones.}
\hyperref[tab:analysis_trans]{Table~\ref*{tab:analysis_trans}} shows that CoCA exceeds RaR on every internalization transition for all three backbones.
The gains are particularly pronounced on Qwen3-8B, where \textit{Learnable} ${\color{mydarkred}\Longrightarrow}$ \textit{Saturated} and \textit{Discriminative} ${\color{mygreen}\Longrightarrow}$ \textit{Saturated} increase by \textcolor{mygreen}{$+7.06\%$} and \textcolor{mygreen}{$+10.53\%$}, respectively.
CoCA also suppresses degradation: relative to RaR, \textit{Learnable} ${\color{mydarkred}\Longrightarrow}$ \textit{Too-Hard} changes by \textcolor{mygreen}{$-0.51\%$}, \textcolor{mygreen}{$-0.07\%$}, and \textcolor{mygreen}{$-1.11\%$} across Qwen3-4B, Qwen3-8B, and Gemma-3-4B-it, while \textit{Saturated} ${\color{mydarkred}\Longrightarrow}$ \textit{Others} changes by \textcolor{mygreen}{$-0.65\%$} and \textcolor{mygreen}{$-0.83\%$} on Qwen3-4B and Gemma-3-4B-it.
These transition patterns demonstrate that CoCA more effectively internalizes guided capabilities while promoting more stable behavior.

\noindent \textbf{CoCA couples efficient learning with sustained exploration and a higher sampling ceiling.}
For Qwen3-8B, as shown in \hyperref[fig:training_dynamic_and_best_of_k]{Figure~\ref*{fig:training_dynamic_and_best_of_k}}, CoCA maintains higher actor entropy than RaR over most of training, consistent with broader exploration and less premature policy contraction.
At step $250$, CoCA reaches $42.19\%$, slightly exceeding RaR at step $500$ ($41.88\%$) with approximately $2.0\times$ fewer training steps.
Its Best-of-$K$ advantage over RaR further remains between \textcolor{mygreen}{$+3.85\%$} and \textcolor{mygreen}{$+4.45\%$} for $K\in\{1,2,4,8,16\}$.
Overall, these results suggest that CoCA reaches comparable performance in fewer training steps while sustaining exploration and expanding the attainable capability frontier.

\section{Related Work}
\label{sec:related_works}


\noindent \textbf{Rubric as Reward in RL.}  
Open-ended generation~\citep{arora2025healthbench,yifei2026researchqa,liu2025heartbench} lacks deterministic verifiers, making scalar or pairwise rewards too coarse to capture multidimensional response quality~\citep{chen2026holistic}. 
Rubric-based RL instead decomposes evaluation into interpretable criteria, with RaR~\citep{gunjal2026rubrics} establishing this formulation for policy optimization beyond verifiable domains. 
Subsequent work improves different stages of this pipeline: OpenRubrics~\citep{liu2026openrubrics} scales rubric generation from contrastive preferences, Rubric-ARM~\citep{xu2026alternating} jointly optimizes the rubric generator and judge, and RuscaRL~\citep{zhou2026breaking} uses rubric criteria to scaffold policy exploration. 
To overcome static evaluation targets, DR Tulu~\citep{shao2026dr} and Online Rubrics~\citep{rezaei2026online} update criteria from policy-generated responses.
Although effective, these methods construct or update criteria from static annotations or response contrasts without tracking the policy's evolving ability to fulfill the task objective, risking stylistic drift.
In contrast, CoCA adapts criterion difficulty to the policy's evolving capabilities while preserving the original evaluation intent (further verified in  \hyperref[sec:rubric_intent_preservation]{Appendix~\ref*{sec:rubric_intent_preservation}}).



\noindent \textbf{On-Policy Distillation and Self-Distillation.}
On-policy distillation trains a student on its own rollouts under dense teacher supervision, reducing the train--inference mismatch of off-policy distillation~\citep{ICLR2024_5be69a58,gu2024minillm,yang2026self}. OPSD~\citep{zhao2026selfdistilled,wang2026trace} removes the need for a separate teacher by conditioning the same model on privileged information and matching its token distributions along student trajectories. Recent variants instantiate the privileged information with rich environmental feedback~\citep{hubotter2026reinforcement}, binary rewards and self-revision~\citep{he2026selfdistillation}, or explicit rubrics~\citep{rezaei2026rubric}.
Beyond logit matching, ROPD~\citep{fang2026rubric} induces rubrics from teacher--student contrasts to score student rollouts, enabling black-box on-policy distillation without teacher logits.
However, sample-level teacher signals may overlook whether criterion-level gains are elicitable and transferable, risking weak or overly difficult supervision.
We instead transfer capability-qualified frontiers and jointly expand their source, coupling selective distillation with capability-aware rubric calibration.
CriPO~\citep{xia2026enhancing} is a concurrent study mostly related to ours, which combines criterion-directed self-distillation with token-level advantage flipping in GRPO; in contrast, CoCA uses dual-group capability diagnosis to drive closed-loop policy--rubric co-evolution.
Further Comparison is provided in \hyperref[sec:extended_related_work]{Appendix~\ref*{sec:extended_related_work}}.
\section{Conclusion}
\label{sec:conclusion}

In this work, we propose \textbf{CoCA}, a capability-aware framework for jointly evolving rubrics and policies in open-ended generation.
In contrast to prior methods that treat rubric adaptation, guided exploration, and knowledge transfer in isolation, CoCA unifies these processes within a capability-aware policy--rubric co-evolution loop.
This closed loop keeps the optimization objective aligned with the policy's evolving capability frontier while selectively transferring qualified guided gains into query-only behavior.
Extensive experiments demonstrate that CoCA consistently improves performance and out-of-domain generalization across diverse tasks and models.
Further analyses reveal complementary effects among CoCA's components and more favorable capability-state transitions, indicating that guided gains are progressively consolidated into stable query-only behavior. 

\subsection*{AI use statement}


We used LLMs to construct psychology-domain training data from CPsyCoun, including queries and task-specific rubrics.
LLMs also supported rubric rewriting and served as judges for reward computation, capability diagnosis, and evaluation, including the rubric intent-preservation audit.
The data-construction procedures, prompts, and rubric calibration examples are provided in \hyperref[sec:datasets_details]{Appendix~\ref*{sec:datasets_details}}, \hyperref[sec:prompt_design]{Appendix~\ref*{sec:prompt_design}}, and \hyperref[sec:case_study]{Appendix~\ref*{sec:case_study}}, respectively.
In preparing the manuscript, we used LLMs to improve clarity, grammar, and readability.
The authors take responsibility for the research design, experimental results, their interpretation, and the final manuscript.

\subsection*{Ethics statement}
This work studies post-training methods for improving rubric-guided learning in open-ended domains. We use existing publicly available research benchmarks and pretrained models, and do not collect new human-subject data or attempt to identify individuals. Because our experiments include medical and psychological content, model outputs may contain inaccurate, biased, or otherwise harmful advice; the resulting models are intended only for research and should not be used for diagnosis, treatment, crisis intervention, or psychological counseling. Rubric generation and LLM-based judging can inherit biases from source data and evaluator models, while adaptive optimization may amplify misspecified criteria or judge artifacts. Benchmark improvements therefore do not establish real-world reliability or clinical safety. Deployment-oriented use would require domain-expert review, demographic and cultural bias audits, privacy assessment, and application-specific safety evaluation.

\subsection*{Reproducibility statement}
Dataset construction and preprocessing are detailed in \hyperref[sec:datasets_details]{Appendix~\ref*{sec:datasets_details}}, while model configurations, training hyperparameters, and implementation details are documented in \hyperref[sec:implementation_details]{Appendix~\ref*{sec:implementation_details}}. All prompts are provided in \hyperref[sec:prompt_design]{Appendix~\ref*{sec:prompt_design}}. 



\bibliography{iclr2027_conference}

@article{li2026evolm,
  title={EvoLM: Self-Evolving Language Models through Co-Evolved Discriminative Rubrics},
  author={Li, Shuyue Stella and Xin, Rui and Xiao, Teng and Wang, Yike and Shao, Rulin and Hao, Zoey and Sclar, Melanie and Oh, Sewoong and Brahman, Faeze and Koh, Pang Wei and others},
  journal={arXiv preprint arXiv:2605.03871},
  year={2026}
}

@inproceedings{
gunjal2026rubrics,
title={Rubrics as Rewards: Reinforcement Learning Beyond Verifiable Domains},
author={Anisha Gunjal and Anthony Wang and Elaine Lau and Vaskar Nath and Yunzhong He and Bing Liu and Sean M. Hendryx},
booktitle={The Fourteenth International Conference on Learning Representations},
year={2026},
url={https://openreview.net/forum?id=c1bTcrDmt4}
}

@inproceedings{
zhou2026breaking,
title={Breaking the Exploration Bottleneck: Rubric-Scaffolded Reinforcement Learning for Open-Ended {LLM} Reasoning},
author={Yang Zhou and Sunzhu Li and Shunyu Liu and Wenkai Fang and Kongcheng Zhang and Jiale Zhao and Jingwen Yang and Yihe Zhou and Jianwei Lv and Tongya Zheng and Hengtong Lu and Chen Wei and Xie Yan and Mingli Song},
booktitle={Forty-third International Conference on Machine Learning},
year={2026},
url={https://openreview.net/forum?id=IM8ZHjhD0v}
}

@inproceedings{
wang2026textgrad,
title={Text2Grad: Reinforcement Learning from Natural Language Feedback},
author={Hanyang Wang and Lu Wang and Chaoyun Zhang and Tianjun Mao and Si Qin and Qingwei Lin and Saravan Rajmohan and Dongmei Zhang},
booktitle={The Fourteenth International Conference on Learning Representations},
year={2026},
url={https://openreview.net/forum?id=SIE9fNq8lk}
}

@inproceedings{
zhang2026advancing,
title={Advancing {LLM} Reasoning with Natural Language and Numerical Feedback},
author={Xiaoying Zhang and Yipeng Zhang and Hao Sun and Kaituo Feng and Chaochao Lu and Chao Yang and Helen M. Meng},
booktitle={Forty-third International Conference on Machine Learning},
year={2026},
url={https://openreview.net/forum?id=gz7hVnrRWq}
}

@inproceedings{
rezaei2026online,
title={Online Rubrics Elicitation from Pairwise Comparisons},
author={MohammadHossein Rezaei and Robert Vacareanu and Zihao Wang and Clinton Wang and Bing Liu and Yunzhong He and Afra Feyza Aky{\"u}rek},
booktitle={Forty-third International Conference on Machine Learning},
year={2026},
url={https://openreview.net/forum?id=BjypaSs3sS}
}

@article{rezaei2026rubric,
  title={Rubric-guided self-distillation: Post-training without rubric verifiers},
  author={Rezaei, MohammadHossein and Mahmoud, Anas and Wang, Zihao and Tyagi, Utkarsh and Gosai, Advait and Dumitru, Razvan-Gabriel and Sabharwal, Aakash and Liu, Bing and He, Yunzhong},
  journal={arXiv preprint arXiv:2606.12507},
  year={2026}
}

@article{ding2026evorubrics,
  title={EvoRubrics: Dynamic Rubrics as Rewards via Adversarial Co-Evolution for LLM Reinforcement Learning},
  author={Ding, Hongxin and Huang, Baixiang and Fang, Yue and Liao, Weibin and Li, Zheng and Zhang, Jinyang and Wu, Zhijing and Zhao, Junfeng and Wang, Yasha},
  journal={arXiv preprint arXiv:2606.23038},
  year={2026}
}

@inproceedings{
zhao2026selfdistilled,
title={Self-Distilled Reasoner: On-Policy Self-Distillation for Large Language Models},
author={Siyan Zhao and Zhihui Xie and Mengchen Liu and Jing Huang and Guan Pang and Feiyu Chen and Aditya Grover},
booktitle={Forty-third International Conference on Machine Learning},
year={2026},
url={https://openreview.net/forum?id=Jpxfof0EaS}
}

@inproceedings{
hubotter2026reinforcement,
title={Reinforcement Learning via Self-Distillation},
author={Jonas H{\"u}botter and Frederike L{\"u}beck and Lejs Deen Behric and Anton Baumann and Marco Bagatella and Daniel Marta and Ido Hakimi and Idan Shenfeld and Thomas Kleine Buening and Carlos Guestrin and Andreas Krause},
booktitle={Forty-third International Conference on Machine Learning},
year={2026},
url={https://openreview.net/forum?id=QkfkxyRizZ}
}

@inproceedings{
li2026rethinking,
title={Rethinking On-Policy Distillation of Large Language Models: Phenomenology, Mechanism, and Recipe},
author={Yaxuan Li and Yuxin Zuo and Bingxiang He and Jinqian Zhang and Chaojun Xiao and Cheng Qian and Tianyu Yu and Huan-ang Gao and Wenkai Yang and Zhiyuan Liu and Ning Ding},
booktitle={ICML 2026 Workshop on Foundations of Deep Generative Models: Understanding Memorization, Generalization, and Reasoning},
year={2026},
url={https://openreview.net/forum?id=aaDDXXDGXB}
}

@inproceedings{
he2026selfdistillation,
title={Self-Distillation Zero: Self-Revision Turns Binary Rewards into Dense Supervision},
author={Yinghui He and Simran Kaur and Adithya Bhaskar and Yongjin Yang and Jiarui Liu and Narutatsu Ri and Liam H Fowl and Abhishek Panigrahi and Danqi Chen and Sanjeev Arora},
booktitle={ICML 2026 Workshop on Foundations of Deep Generative Models: Understanding Memorization, Generalization, and Reasoning},
year={2026},
url={https://openreview.net/forum?id=Eob8DfeK0J}
}

@article{xia2026enhancing,
  title={Enhancing Rubric-based RL via Self-Distillation},
  author={Xia, Mingxuan and Yang, Yuhang and Ye, Chao and Zhu, Shuai and Yang, Shenzhi and Zhu, Guangcheng and Zhang, Yuhang and Peng, Cheng and Wang, Haobo and Wang, Siqing},
  journal={arXiv preprint arXiv:2607.18082},
  year={2026}
}

@article{liu2026crosslingual,
  title={Crosslingual on-policy self-distillation for multilingual reasoning},
  author={Liu, Yihong and Zhao, Raoyuan and Hedderich, Michael A and Sch{\"u}tze, Hinrich},
  journal={arXiv preprint arXiv:2605.09548},
  year={2026}
}

@article{han2026adaptive,
  title={Adaptive Teacher Exposure for Self-Distillation in LLM Reasoning},
  author={Han, Zihao and Zhang, Tiangang and Wang, Huaibin and Sun, Yilun},
  journal={arXiv preprint arXiv:2605.11458},
  year={2026}
}

@article{fang2026rubric,
  title={Rubric-based on-policy distillation},
  author={Fang, Junfeng and Hong, Zhepei and Zheng, Mao and Song, Mingyang and Li, Gengsheng and Jiang, Houcheng and Zhang, Dan and Guo, Haiyun and Wang, Xiang and Chua, Tat-Seng},
  journal={arXiv preprint arXiv:2605.07396},
  year={2026}
}

@inproceedings{bengio2009curriculum,
  title={Curriculum learning},
  author={Bengio, Yoshua and Louradour, J{\'e}r{\^o}me and Collobert, Ronan and Weston, Jason},
  booktitle={Proceedings of the 26th annual international conference on machine learning},
  pages={41--48},
  year={2009}
}

@article{narvekar2020curriculum,
  title={Curriculum learning for reinforcement learning domains: A framework and survey},
  author={Narvekar, Sanmit and Peng, Bei and Leonetti, Matteo and Sinapov, Jivko and Taylor, Matthew E and Stone, Peter},
  journal={Journal of Machine Learning Research},
  volume={21},
  number={181},
  pages={1--50},
  year={2020}
}

@article{yang2025qwen3,
  title={Qwen3 technical report},
  author={Yang, An and Li, Anfeng and Yang, Baosong and Zhang, Beichen and Hui, Binyuan and Zheng, Bo and Yu, Bowen and Gao, Chang and Huang, Chengen and Lv, Chenxu and others},
  journal={arXiv preprint arXiv:2505.09388},
  year={2025}
}

@article{team2025gemma,
  title={Gemma 3 technical report},
  author={Team, Gemma and Kamath, Aishwarya and Ferret, Johan and Pathak, Shreya and Vieillard, Nino and Merhej, Ramona and Perrin, Sarah and Matejovicova, Tatiana and Ram{\'e}, Alexandre and Rivi{\`e}re, Morgane and others},
  journal={arXiv preprint arXiv:2503.19786},
  year={2025}
}

@article{arora2025healthbench,
  title={Healthbench: Evaluating large language models towards improved human health},
  author={Arora, Rahul K and Wei, Jason and Hicks, Rebecca Soskin and Bowman, Preston and Qui{\~n}onero-Candela, Joaquin and Tsimpourlas, Foivos and Sharman, Michael and Shah, Meghan and Vallone, Andrea and Beutel, Alex and others},
  journal={arXiv preprint arXiv:2505.08775},
  year={2025}
}

@inproceedings{zhang-etal-2025-llmeval,
    title = "{LLME}val-{M}ed: A Real-world Clinical Benchmark for Medical {LLM}s with Physician Validation",
    author = "Zhang, Ming  and
      Shen, Yujiong  and
      Li, Zelin  and
      Sha, Huayu  and
      Hu, Binze  and
      Wang, Yuhui  and
      Huang, Chenhao  and
      Liu, Shichun  and
      Tong, Jingqi  and
      Jiang, Changhao  and
      Chai, Mingxu  and
      Xi, Zhiheng  and
      Dou, Shihan  and
      Gui, Tao  and
      Zhang, Qi  and
      Huang, Xuanjing",
    editor = "Christodoulopoulos, Christos  and
      Chakraborty, Tanmoy  and
      Rose, Carolyn  and
      Peng, Violet",
    booktitle = "Findings of the Association for Computational Linguistics: EMNLP 2025",
    month = nov,
    year = "2025",
    address = "Suzhou, China",
    publisher = "Association for Computational Linguistics",
    url = "https://aclanthology.org/2025.findings-emnlp.263/",
    doi = "10.18653/v1/2025.findings-emnlp.263",
    pages = "4888--4914",
    ISBN = "979-8-89176-335-7",
}

@article{yifei2026researchqa,
  title={Researchqa: Evaluating scholarly question answering at scale across 75 fields with survey-mined questions and rubrics},
  author={Yifei, Li S and Chang, Allen and Malaviya, Chaitanya and Yatskar, Mark},
  journal={Transactions of the Association for Computational Linguistics},
  volume={14},
  pages={1344--1368},
  year={2026},
  publisher={MIT Press 255 Main Street, 9th Floor, Cambridge, Massachusetts 02142, USA~…}
}

@inproceedings{zhang2024cpsycoun,
  title={Cpsycoun: A report-based multi-turn dialogue reconstruction and evaluation framework for chinese psychological counseling},
  author={Zhang, Chenhao and Li, Renhao and Tan, Minghuan and Yang, Min and Zhu, Jingwei and Yang, Di and Zhao, Jiahao and Ye, Guancheng and Li, Chengming and Hu, Xiping},
  booktitle={Findings of the Association for Computational Linguistics: ACL 2024},
  pages={13947--13966},
  year={2024}
}

@article{liu2025heartbench,
  title={HeartBench: Probing Core Dimensions of Anthropomorphic Intelligence in LLMs},
  author={Liu, Jiaxin and Tu, Peiyi and Chen, Wenyu and Zhuang, Yihong and Ling, Xinxia and Zhou, Anji and Wang, Chenxi and Han, Zhuo and Yang, Zhengkai and Zhao, Junbo and others},
  journal={arXiv preprint arXiv:2512.21849},
  year={2025}
}

@article{liu2025deepseek,
  title={Deepseek-v3.2: Pushing the frontier of open large language models},
  author={Liu, Aixin and Mei, Aoxue and Lin, Bangcai and Xue, Bing and Wang, Bingxuan and Xu, Bingzheng and Wu, Bochao and Zhang, Bowei and Lin, Chaofan and Dong, Chen and others},
  journal={arXiv preprint arXiv:2512.02556},
  year={2025}
}

@inproceedings{liusie-etal-2024-llm,
    title = "{LLM} Comparative Assessment: Zero-shot {NLG} Evaluation through Pairwise Comparisons using Large Language Models",
    author = "Liusie, Adian  and
      Manakul, Potsawee  and
      Gales, Mark",
    editor = "Graham, Yvette  and
      Purver, Matthew",
    booktitle = "Proceedings of the 18th Conference of the European Chapter of the Association for Computational Linguistics (Volume 1: Long Papers)",
    month = mar,
    year = "2024",
    address = "St. Julian{'}s, Malta",
    publisher = "Association for Computational Linguistics",
    url = "https://aclanthology.org/2024.eacl-long.8/",
    doi = "10.18653/v1/2024.eacl-long.8",
    pages = "139--151",
}

@inproceedings{qin2024infobench,
  title={Infobench: Evaluating instruction following ability in large language models},
  author={Qin, Yiwei and Song, Kaiqiang and Hu, Yebowen and Yao, Wenlin and Cho, Sangwoo and Wang, Xiaoyang and Wu, Xuansheng and Liu, Fei and Liu, Pengfei and Yu, Dong},
  booktitle={Findings of the Association for Computational Linguistics: ACL 2024},
  pages={13025--13048},
  year={2024}
}

@inproceedings{jiang2024followbench,
  title={Followbench: A multi-level fine-grained constraints following benchmark for large language models},
  author={Jiang, Yuxin and Wang, Yufei and Zeng, Xingshan and Zhong, Wanjun and Li, Liangyou and Mi, Fei and Shang, Lifeng and Jiang, Xin and Liu, Qun and Wang, Wei},
  booktitle={Proceedings of the 62nd Annual Meeting of the Association for Computational Linguistics (Volume 1: Long Papers)},
  pages={4667--4688},
  year={2024}
}

@inproceedings{he2026advancedif,
  title={Advancedif: Rubric-based benchmarking and reinforcement learning for advancing llm instruction following},
  author={He, Yun and Li, Wenzhe and Zhang, Hejia and Li, Songlin and Mandyam, Karishma and Khosla, Sopan and Xiong, Yuanhao and Wang, Nanshu and Peng, Xiaoliang and Li, Beibin and others},
  booktitle={Proceedings of the 64th Annual Meeting of the Association for Computational Linguistics (Volume 1: Long Papers)},
  pages={18003--18022},
  year={2026}
}

@inproceedings{liu2026openrubrics,
  title={Openrubrics: Towards scalable synthetic rubric generation for reward modeling and llm alignment},
  author={Liu, Tianci and Xu, Ran and Yu, Tony and Hong, Ilgee and Yang, Carl and Zhao, Tuo and Wang, Haoyu},
  booktitle={Proceedings of the 64th Annual Meeting of the Association for Computational Linguistics (Volume 1: Long Papers)},
  pages={17417--17437},
  year={2026}
}

@inproceedings{
xu2026alternating,
title={Alternating Reinforcement Learning for Rubric-Based Reward Modeling in Non-Verifiable {LLM} Post-Training},
author={Ran Xu and Tianci Liu and Zihan Dong and Tony Yu and Ilgee Hong and Carl Yang and Linjun Zhang and Tuo Zhao and Haoyu Wang},
booktitle={Forty-third International Conference on Machine Learning},
year={2026},
url={https://openreview.net/forum?id=3SDg8dtkS8}
}

@inproceedings{
shao2026dr,
title={{DR} Tulu: Reinforcement Learning with Evolving Rubrics for Deep Research},
author={Rulin Shao and Akari Asai and Shannon Zejiang Shen and Hamish Ivison and Varsha Kishore and Jingming Zhuo and Xinran Zhao and Molly Park and Samuel G. Finlayson and David Sontag and Tyler Murray and Sewon Min and Pradeep Dasigi and Luca Soldaini and Faeze Brahman and Wen-tau Yih and Tongshuang Wu and Luke Zettlemoyer and Yoon Kim and Hannaneh Hajishirzi and Pang Wei Koh},
booktitle={Forty-third International Conference on Machine Learning},
year={2026},
url={https://openreview.net/forum?id=97NEP1pyS3}
}

@article{wang2026co,
  title={Co-Evolving LLM Evaluators and Policies via DynamicRubric},
  author={Wang, Beining and Su, Weihang and Tian, Hongtao and Kong, Hao and Yang, Tao and Yao, Ting and Pan, Qingyi and Wu, Yueyue and Ai, Qingyao and Zhang, Min and others},
  journal={arXiv preprint arXiv:2607.20083},
  year={2026}
}

@article{guan2026evorubric,
  title={EvoRubric: Self-Evolving Rubric-Driven RL for Open-Ended Generation},
  author={Guan, Xin and Hu, Xiaomeng and Huang, Shen and Wang, Zhenyi and Zhang, Bo and Li, Zijian and Xie, Pengjun and Liu, Bo and Cao, Jiuxin},
  journal={arXiv preprint arXiv:2605.29847},
  year={2026}
}

@inproceedings{ICLR2024_5be69a58,
 author = {Agarwal, Rishabh and Vieillard, Nino and Zhou, Yongchao and Stanczyk, Piotr and Ramos Garea, Sabela and Geist, Matthieu and Bachem, Olivier},
 booktitle = {International Conference on Learning Representations},
 editor = {B. Kim and Y. Yue and S. Chaudhuri and K. Fragkiadaki and M. Khan and Y. Sun},
 pages = {21246--21263},
 title = {On-Policy Distillation of Language Models: Learning from Self-Generated Mistakes},
 url = {https://proceedings.iclr.cc/paper_files/paper/2024/file/5be69a584901a26c521c2b51e40a4c20-Paper-Conference.pdf},
 volume = {2024},
 year = {2024}
}

@article{kim2026llm,
  title={LLM-as-a-Tutor: Policy-Aware Prompt Adaptation for Non-Verifiable RL},
  author={Kim, Yujin and Ho, Namgyu and Hwang, Sangmin and Kim, Joonkee and Yang, Yongjin and Bae, Sangmin and Kim, Seungone and Jung, Jaehun and Yun, Se-Young and Song, Hwanjun},
  journal={arXiv preprint arXiv:2607.04412},
  year={2026}
}

@article{chen2026holistic,
  title={From Holistic Evaluation to Structured Criteria: Rubrics Across the Evolving LLM Landscape},
  author={Chen, Hao and Han, Ziyu and Yan, Yukun and Zhu, Qingfu and Sun, Maosong and Che, Wanxiang},
  journal={arXiv preprint arXiv:2606.08625},
  year={2026}
}

@inproceedings{gu2024minillm,
  title={Minillm: Knowledge distillation of large language models},
  author={Gu, Yuxian and Dong, Li and Wei, Furu and Huang, Minlie},
  booktitle={International Conference on Learning Representations},
  volume={2024},
  pages={32694--32717},
  year={2024}
}

@article{yang2026self,
  title={Self-distilled rlvr},
  author={Yang, Chenxu and Qin, Chuanyu and Si, Qingyi and Chen, Minghui and Gu, Naibin and Yao, Dingyu and Lin, Zheng and Wang, Weiping and Wang, Jiaqi and Duan, Nan},
  journal={arXiv preprint arXiv:2604.03128},
  year={2026}
}

@article{wang2026trace,
  title={Trace: Distilling where it matters via token-routed self on-policy alignment},
  author={Wang, Jiaxuan and Ouyang, Xuan and Chen, Zhiyu and Hu, Yulan and Pan, Zheng and Li, Xin and Guo, Lan-Zhe},
  journal={arXiv preprint arXiv:2605.10194},
  year={2026}
}

@article{kim2026does,
  title={Why Does Self-Distillation (Sometimes) Degrade the Reasoning Capability of LLMs?},
  author={Kim, Jeonghye and Luo, Xufang and Kim, Minbeom and Lee, Sangmook and Kim, Dohyung and Jeon, Jiwon and Li, Dongsheng and Yang, Yuqing},
  journal={arXiv preprint arXiv:2603.24472},
  year={2026}
}

@article{guo2025deepseek,
  title={Deepseek-r1: Incentivizing reasoning capability in llms via reinforcement learning},
  author={Guo, Daya and Yang, Dejian and Zhang, Haowei and Song, Junxiao and Wang, Peiyi and Zhu, Qihao and Xu, Runxin and Zhang, Ruoyu and Ma, Shirong and Bi, Xiao and others},
  journal={arXiv preprint arXiv:2501.12948},
  year={2025}
}

@article{shao2024deepseekmath,
  title={Deepseekmath: Pushing the limits of mathematical reasoning in open language models},
  author={Shao, Zhihong and Wang, Peiyi and Zhu, Qihao and Xu, Runxin and Song, Junxiao and Bi, Xiao and Zhang, Haowei and Zhang, Mingchuan and Li, YK and Wu, Yang and others},
  journal={arXiv preprint arXiv:2402.03300},
  year={2024}
}

@inproceedings{su-etal-2026-crossing,
    title = "Crossing the Reward Bridge: Expanding Reinforcement Learning with Verifiable Rewards Across Diverse Domains",
    author = "Su, Yi  and
      Yu, Dian  and
      Song, Linfeng  and
      Li, Juntao  and
      Mi, Haitao  and
      Tu, Zhaopeng  and
      Zhang, Min  and
      Yu, Dong",
    editor = "Liakata, Maria  and
      Moreira, Viviane P.  and
      Zhang, Jiajun  and
      Jurgens, David",
    booktitle = "Proceedings of the 64th Annual Meeting of the {A}ssociation for {C}omputational {L}inguistics (Volume 1: Long Papers)",
    month = jul,
    year = "2026",
    address = "San Diego, California, United States",
    publisher = "Association for Computational Linguistics",
    url = "https://aclanthology.org/2026.acl-long.178/",
    doi = "10.18653/v1/2026.acl-long.178",
    pages = "3872--3892",
    ISBN = "979-8-89176-390-6",
}

@article{huang2025reinforcement,
  title={Reinforcement learning with rubric anchors},
  author={Huang, Zenan and Zhuang, Yihong and Lu, Guoshan and Qin, Zeyu and Xu, Haokai and Zhao, Tianyu and Peng, Ru and Hu, Jiaqi and Shen, Zhanming and Hu, Xiaomeng and others},
  journal={arXiv preprint arXiv:2508.12790},
  year={2025}
}

@article{shen2026rethinking,
  title={Rethinking rubric generation for improving llm judge and reward modeling for open-ended tasks},
  author={Shen, William F and Qiu, Xinchi and Whitehouse, Chenxi and Alazraki, Lisa and Goel, Shashwat and Barbieri, Francesco and Willi, Timon and Mathur, Akhil and Leontiadis, Ilias},
  journal={arXiv preprint arXiv:2602.05125},
  year={2026}
}

@article{yang2026rubrics,
  title={Rubrics on Trial: Evolving Rubrics from a Single Query via Synthetic Pairwise Evidence},
  author={Yang, Haocheng and Pan, Licheng and Li, Xiaoxi and Chen, Zhichao and Zhang, Zhiheng and Lu, Yuan and Li, Haoxuan and Wang, Hao},
  journal={arXiv preprint arXiv:2607.15092},
  year={2026}
}

@misc{zheng2025groupsequencepolicyoptimization,
      title={Group Sequence Policy Optimization}, 
      author={Chujie Zheng and Shixuan Liu and Mingze Li and Xiong-Hui Chen and Bowen Yu and Chang Gao and Kai Dang and Yuqiong Liu and Rui Men and An Yang and Jingren Zhou and Junyang Lin},
      year={2025},
      eprint={2507.18071},
      archivePrefix={arXiv},
      primaryClass={cs.LG},
      url={https://arxiv.org/abs/2507.18071}, 
}

@article{yu2026dapo,
  title={Dapo: An open-source llm reinforcement learning system at scale},
  author={Yu, Qiying and Zhang, Zheng and Zhu, Ruofei and Yuan, Yufeng and Zuo, Xiaochen and Yue, Yu and Dai, Weinan and Fan, Tiantian and Liu, Gaohong and Liu, Lingjun and others},
  journal={Advances in Neural Information Processing Systems},
  volume={38},
  pages={113222--113244},
  year={2026}
}

@article{chen2025minimax,
  title={Minimax-m1: Scaling test-time compute efficiently with lightning attention},
  author={Chen, Aili and Li, Aonian and Gong, Bangwei and Jiang, Binyang and Fei, Bo and Yang, Bo and Shan, Boji and Yu, Changqing and Wang, Chao and Zhu, Cheng and others},
  journal={arXiv preprint arXiv:2506.13585},
  year={2025}
}

@inproceedings{dao2024flashattention,
  title={Flashattention-2: Faster attention with better parallelism and work partitioning},
  author={Dao, Tri},
  booktitle={International Conference on Learning Representations},
  volume={2024},
  pages={35549--35562},
  year={2024}
}

@inproceedings{ShengVERL,
author = {Sheng, Guangming and Zhang, Chi and Ye, Zilingfeng and Wu, Xibin and Zhang, Wang and Zhang, Ru and Peng, Yanghua and Lin, Haibin and Wu, Chuan},
title = {HybridFlow: A Flexible and Efficient RLHF Framework},
year = {2025},
isbn = {9798400711961},
publisher = {Association for Computing Machinery},
address = {New York, NY, USA},
url = {https://doi.org/10.1145/3689031.3696075},
doi = {10.1145/3689031.3696075},
booktitle = {Proceedings of the Twentieth European Conference on Computer Systems},
pages = {1279–1297},
numpages = {19},
location = {Rotterdam, Netherlands},
series = {EuroSys '25}
}

@misc{qwen37,
    title = {{Qwen3.7}: The Agent Frontier},
    url = {https://qwen.ai/blog?id=qwen3.7},
    author = {{Qwen Team}},
    month = {May},
    year = {2026}
}

@article{team2026kimi,
  title={Kimi k2.5: Visual agentic intelligence},
  author={Team, Kimi and Bai, Tongtong and Bai, Yifan and Bao, Yiping and Cai, SH and Cao, Yuan and Chai, Ziwei and Charles, Y and Che, HS and Chen, Cheng and others},
  journal={arXiv preprint arXiv:2602.02276},
  year={2026}
}

@article{zeng2026glm,
  title={Glm-5: from vibe coding to agentic engineering},
  author={Zeng, Aohan and Lv, Xin and Hou, Zhenyu and Du, Zhengxiao and Zheng, Qinkai and Chen, Bin and Yin, Da and Ge, Chendi and Huang, Chenghua and Xie, Chengxing and others},
  journal={arXiv preprint arXiv:2602.15763},
  year={2026}
}

@inproceedings{gao2023scaling,
  title={Scaling laws for reward model overoptimization},
  author={Gao, Leo and Schulman, John and Hilton, Jacob},
  booktitle={International conference on machine learning},
  pages={10835--10866},
  year={2023},
  organization={PMLR}
}

@article{skalse2022defining,
  title={Defining and characterizing reward gaming},
  author={Skalse, Joar and Howe, Nikolaus and Krasheninnikov, Dmitrii and Krueger, David},
  journal={Advances in neural information processing systems},
  volume={35},
  pages={9460--9471},
  year={2022}
}

@InProceedings{pmlr-v306-zhang26fi,
  title = 	 {Interpreting and Enhancing Emotional Circuits in Large Vision-Language Models via Cross-Modal Information Flow},
  author =       {Zhang, Chengsheng and Sun, Chenghao and Xie, Zhining and Tian, Xinmei},
  booktitle = 	 {Proceedings of the 43rd International Conference on Machine Learning},
  pages = 	 {158457--158476},
  year = 	 {2026},
  editor = 	 {Zhang, Tong and Dudik, Miroslav and Jaggi, Martin and Agarwal, Alekh and Li, Sharon and Schuurmans, Dale and Zhu, Jerry and Berkenkamp, Felix and Dong, Hanze and Bietti, Alberto},
  volume = 	 {306},
  series = 	 {Proceedings of Machine Learning Research},
  month = 	 {06--11 Jul},
  publisher =    {PMLR},
}
\bibliographystyle{iclr2027_conference}

\clearpage
\appendix

\addtocontents{toc}{\protect\setcounter{tocdepth}{2}}
\begingroup
    \edef\appendixsavedtocdepth{\number\value{tocdepth}}
    \renewcommand{\contentsname}{Appendix Contents}
    \setcounter{tocdepth}{-1}
    \small
    \tableofcontents
    \setcounter{tocdepth}{\appendixsavedtocdepth}
\endgroup
\clearpage

\begin{breakablealgorithm}
    \caption{CoCA: Co-Evolving Policy and Rubric via Capability Awareness}
    \label{alg:coca}
    \small
    \begin{algorithmic}[1]
        \REQUIRE Training set $\mathcal{D}=\{(x,\mathcal{R}_x)\}$ with $\mathcal{R}_x=\{(c_j,w_j)\}_{j=1}^{J_x}$; policy $\pi_\theta$; fixed judge $\mathcal{J}_\phi$
        \REQUIRE Rollout group sizes $K_{\mathrm{N}},K_{\mathrm{G}}$; frontier margin $\delta\geq0$; loss coefficients $\alpha$ (Frontier-OPSD) and $\beta$ (Guided GRPO); retained-token count $k_{\mathrm{tok}}$
        \ENSURE Optimized query-only policy $\pi_\theta$
        \REPEAT
            \STATE Set $\theta_{\mathrm{old}}\leftarrow\theta$, sample a mini-batch $\mathcal{B}\subset\mathcal{D}$, and initialize $\mathcal{L}_{\mathcal{B}}\leftarrow 0$
            \FOR{$(x,\mathcal{R}_x)\in\mathcal{B}$}
                \STATE \textcolor{cocapurple}{$\blacktriangleright$\hspace{0.35em}\textbf{\textit{Capability Probing}}}~(Sec.~\ref{sec:capability_probing})
                \STATE Independently sample $\mathcal{Y}_x^{\mathrm{N}}$ ($K_{\mathrm{N}}$ responses) and $\mathcal{Y}_x^{\mathrm{G}}$ ($K_{\mathrm{G}}$ responses) via Eq.~\ref{eq:dual_group_rollouts}
                \STATE Evaluate $z_{i,j}^{v}=\mathcal{J}_\phi(x,y_i^v,c_j)$ for $v\in\{\mathrm{N},\mathrm{G}\}$, $i=1,\ldots,K_v$, and $j=1,\ldots,J_x$
                \STATE Compute $s_{i,j}^{v}$ and $\bar{s}_j^{v}$ via Eq.~\ref{eq:target_achievement}
                \STATE Diagnose each $c_j$ from $(\bar{s}_j^{\mathrm{N}},\bar{s}_j^{\mathrm{G}})$ via Eq.~\ref{eq:capability_state}, yielding $d_j$
                \STATE \textcolor{cocapurple}{$\blacktriangleright$\hspace{0.35em}\textbf{\textit{Rubric Calibration}}}~(Sec.~\ref{sec:rubric_calibration})
                \FOR{$j=1,\ldots,J_x$}
                    \IF{$d_j=\textit{Saturated}$}
                        \STATE $c'_j\leftarrow\operatorname{Strengthen}(x,c_j,w_j,\mathcal{Y}_x^{\mathrm{N}},\mathcal{Y}_x^{\mathrm{G}})$
                    \ELSIF{$d_j=\textit{Too-Hard}$}
                        \STATE $c'_j\leftarrow\operatorname{Weaken}(x,c_j,w_j,\mathcal{Y}_x^{\mathrm{N}},\mathcal{Y}_x^{\mathrm{G}})$
                    \ELSE
                        \STATE $c'_j\leftarrow c_j$
                    \ENDIF
                \ENDFOR
                \STATE Form $\mathcal{R}'_x=\{(c'_j,w_j)\}_{j=1}^{J_x}$; obtain $\tilde{z}_{i,j}^{v}$ by selectively rejudging modified pairs via Eq.~\ref{eq:selective_rejudging}
                \STATE Compute $\tilde{s}_{i,j}^{v}$ using $\tilde{z}_{i,j}^{v}$ in Eq.~\ref{eq:target_achievement}, and $\tilde{r}_i^{v}$ via Eq.~\ref{eq:reconstructed_reward}
                \STATE \textcolor{cocapurple}{$\blacktriangleright$\hspace{0.35em}\textbf{\textit{Frontier Selection and Gating}}}~(Sec.~\ref{sec:frontier_opsd})
                \STATE Set $\mathcal{I}_x^{\mathrm{L}}\leftarrow\{j\in\{1,\ldots,J_x\}:d_j=\textit{Learnable}\}$ and $\mathcal{R}_x^{\mathrm{L}}\leftarrow\{(c'_j,w_j):j\in\mathcal{I}_x^{\mathrm{L}}\}$
                \STATE Select $i_x^{\mathrm{C}},y_x^{\mathrm{C}}$ via Eq.~\ref{eq:current_frontier}; construct $\mathcal{E}_x^{\mathrm{G}}$ via Eq.~\ref{eq:guided_eligible_set}
                \STATE Initialize $g(x)\leftarrow 0$
                \IF{$\mathcal{I}_x^{\mathrm{L}}\neq\emptyset$ and $\mathcal{E}_x^{\mathrm{G}}\neq\emptyset$}
                    \STATE Select $i_x^{\mathrm{G}},y_x^{\mathrm{G}}$ via Eq.~\ref{eq:guided_frontier}
                    \STATE Set $\Delta_x\leftarrow\tilde{r}_{i_x^{\mathrm{G}}}^{\mathrm{G}}-\tilde{r}_{i_x^{\mathrm{C}}}^{\mathrm{N}}$ and $g(x)\leftarrow\mathbbm{1}[\Delta_x>\delta]$
                \ENDIF
                \STATE \textcolor{cocapurple}{$\blacktriangleright$\hspace{0.35em}\textbf{\textit{Joint Optimization}}}~(Sec.~\ref{sec:joint_optimization})
                \STATE Compute $\widehat{A}_i^v$ separately within each group via Eq.~\ref{eq:groupwise_advantages}
                \STATE Compute $\mathcal{L}_{\mathrm{GRPO}}^{\mathrm{N}}(\theta;x)$ over all $K_{\mathrm{N}}$ Normal responses with context $x$
                \IF{$g(x)=1$}
                    \STATE Evaluate $p_t^{\mathrm{S}}$ and detached $p_t^{\mathrm{T}}$ on prefixes of $y_x^{\mathrm{C}}$ via Eq.~\ref{eq:frontier_teacher_student}
                    \STATE Compute $\mathcal{L}_{\mathrm{Frontier\text{-}OPSD}}(\theta;x)$ via Eq.~\ref{eq:frontier_opsd_loss}, using the teacher's top-$k_{\mathrm{tok}}$ tokens plus one tail-mass bin
                    \STATE $\mathcal{L}_x\leftarrow\mathcal{L}_{\mathrm{GRPO}}^{\mathrm{N}}(\theta;x)+\alpha\mathcal{L}_{\mathrm{Frontier\text{-}OPSD}}(\theta;x)$
                \ELSE
                    \STATE Compute $\mathcal{L}_{\mathrm{GRPO}}^{\mathrm{G}}(\theta;x)$ over all $K_{\mathrm{G}}$ Guided responses with their original context $(x,\mathcal{R}_x)$
                    \STATE $\mathcal{L}_x\leftarrow\mathcal{L}_{\mathrm{GRPO}}^{\mathrm{N}}(\theta;x)+\beta\mathcal{L}_{\mathrm{GRPO}}^{\mathrm{G}}(\theta;x)$
                \ENDIF
                \STATE $\mathcal{L}_{\mathcal{B}}\leftarrow\mathcal{L}_{\mathcal{B}}+\mathcal{L}_x$
            \ENDFOR
            \STATE $\theta\leftarrow\operatorname{OptimizerStep}\!\left(\theta,\nabla_\theta\mathcal{L}_{\mathcal{B}}/|\mathcal{B}|\right)$
        \UNTIL{the training budget is exhausted}
    \end{algorithmic}
\end{breakablealgorithm}

\section{Extended Related Work and Comparison}
\label{sec:extended_related_work}

This section extends Section~\ref{sec:related_works} with a mechanism-level comparison of representative rubric-based methods (\hyperref[tab:related_method_comparison]{Table~\ref*{tab:related_method_comparison}}).
The comparison separates available learning mechanisms from the feedback-dependent decisions that coordinate them.

\vspace{3pt}
\begin{table}[!htbp]
    \centering
    \caption{Comparison of representative rubric-based methods.
        Adaptive rubrics respond to policy feedback.
        Token-level signals denote rubric-informed supervision beyond scalar advantages. 
        State-based learning uses criterion diagnostics to select policy-update interventions.
    }
    \label{tab:related_method_comparison}
    \small
    \setlength{\tabcolsep}{3pt}
    \renewcommand{\arraystretch}{1.10}
    \resizebox{\linewidth}{!}{%
    \begin{tabular}{@{}lcccc}
        \toprule
        \multirow[c]{3}{*}{\textbf{Method}}
        & \multicolumn{3}{c}{\textbf{Learning Mechanisms}}
        & \multirow[c]{3}{*}{\textbf{Decision Coupling}} \\
        \cmidrule(lr){2-4}
        & \makecell{\textbf{Adaptive}\\\textbf{Rubrics}}
        & \makecell{\textbf{Token-level}\\\textbf{Signals}}
        & \makecell{\textbf{State-based}\\\textbf{Learning}}
        & \\
        \midrule
        RaR~\citep{gunjal2026rubrics}
        & $\times$ & $\times$ & $\times$
        & Fixed Rubric $\to$ Policy Rewards \\
        \addlinespace[3pt]
        RuscaRL~\citep{zhou2026breaking}
        & $\times$ & $\times$ & $\times$
        & Scheduled Rubric Guidance \\
        \addlinespace[3pt]
        Online Rubrics~\citep{rezaei2026online}
        & $\checkmark$ & $\times$ & $\times$
        & Response Contrasts $\to$ Criteria \\
        \addlinespace[3pt]
        RGSD~\citep{rezaei2026rubric}
        & $\times$ & $\checkmark$ & $\times$
        & Fixed Rubric $\to$ Teacher Signals \\
        \addlinespace[3pt]
        CriPO~\citep{xia2026enhancing}
        & $\times$ & $\checkmark$ & $\checkmark$
        & Criterion States $\to$ Injection and Advantage Flipping \\
        \addlinespace[3pt]
        EvoRubric~\citep{guan2026evorubric}
        & $\checkmark$ & $\times$ & $\times$
        & Shared-Policy Co-Training \\
        \addlinespace[3pt]
        EvoRubrics~\citep{ding2026evorubrics}
        & $\checkmark$ & $\times$ & $\times$
        & Revision-Driven Co-Training \\
        \addlinespace[3pt]
        DynamicRubric~\citep{wang2026co}
        & $\checkmark$ & $\times$ & $\times$
        & Alternating Rubric/Policy Training \\
        \midrule
        \rowcolor{mypurple}
        \textbf{CoCA (Ours)}
        & $\checkmark$ & $\checkmark$ & $\checkmark$
        & Shared Capability Diagnosis $\to$ Calibration and Learning \\
        \bottomrule
    \end{tabular}%
    }
\vspace{3pt}

\end{table}

\noindent \textbf{Capability diagnosis as a shared control signal.} 
EvoRubrics and DynamicRubric co-evolve reward construction with policy learning~\citep{ding2026evorubrics,wang2026co}, and CriPO uses criterion states to direct distillation and advantage correction~\citep{xia2026enhancing}.
CoCA instead derives a shared control signal by contrasting criterion achievement between query-only and rubric-conditioned rollouts of the same policy.
The resulting diagnosis simultaneously determines which criteria to recalibrate and which guidance-elicited targets a frontier must satisfy.
Crucially, this separates behaviors that guidance can elicit from targets unmet in both sampled groups, rather than reading query-only failure as evidence of absent capability.
Frontier eligibility and the full-rubric reward advantage then determine the choice between self-distillation and guided optimization, with \textit{Normal} GRPO active throughout (Eqs.~\ref{eq:capability_state}, \ref{eq:frontier_gate}, and~\ref{eq:joint_objective}).
As the policy evolves, renewed diagnosis updates both calibration decisions and learning targets, linking reward adaptation, capability expansion, and internalization in a closed feedback loop.

\section{CoCa Algorithm}
\label{sec:algorithm}

Algorithm~\ref{alg:coca} summarizes the complete CoCa training loop. Each iteration uses dual-group rollouts to diagnose policy capability, calibrates the rubric against the resulting empirical frontier, and then routes each query to either frontier transfer or frontier expansion while retaining \textit{Normal} GRPO.
Notably, the rewritten criteria serve only to reconstruct rewards for the current rollout groups and are not carried over to subsequent iterations; each revisit to a query starts from its original rubric.

\section{Detailed experimental settings}

\subsection{Open-ended Generation Datasets}
\label{sec:datasets_details}

Our experiments cover medical, scientific, and psychological generation, together with out-of-domain instruction following; dataset statistics are summarized in \hyperref[dataset_details]{Table~\ref*{dataset_details}}.
For datasets evaluated with \textit{Rubric Score}, we aggregate criterion-level judgments using Eq.~\ref{eq:rubric_reward_detailed} and average across test examples; all metrics are reported on a $0$--$100$ scale.

\begin{table*}[h]
    \centering
    \caption{
        Statistics of the datasets used in our experiments.
        ``\# Avg. Criteria'' counts rubric items or instruction constraints per sample.
    }
    \label{dataset_details}
    \vspace{-5pt}
    \resizebox{\linewidth}{!}{
        \begin{tabular}{llcccl}
            \whline
            \rowcolor{mygray}
            \textbf{Domain} & \textbf{Dataset} & \textbf{Split} & \textbf{\# Samples} & \textbf{\# Avg. Criteria} & \textbf{Metric} \\
            \hline

            \multirow{3}{*}{\textbf{Medical Domain}}
             & HealthBench & Train/Test & $4,000/1,000$ & $11.35/11.85$ & Rubric Score \\
             & RaR-Medicine & Test & $2,242$ & $9.59$ & Rubric Score \\
             & LLMEval-Med & Test & $667$ & -- & Overall Performance (OP) \\
            \hline

            \multirow{2}{*}{\textbf{Science Domain}}
             & ResearchQA & Train/Valid & $4,000/500$ & $7.53/7.46$ & Rubric Score \\
             & RaR-Science & Test & $500$ & $7.46$ & Rubric Score \\
            \hline

            \multirow{2}{*}{\textbf{Psychological Domain}}
             & CPsyCoun & Train/Test & $2,400/500$ & $8.48/8.46$ & Rubric Score \\
             & HeartBench & Test & $296$ & $10.52$ & Overall Score \\
            \hline

            \multirow{3}{*}{\shortstack{\textbf{Instruction Following} \\ \textbf{(OOD)}}}
             & InfoBench & Test & $500$ & $4.50$ & Requirement Pass Rate \\
             & FollowBench & Test & $820$ & $3.00$ & Hard Satisfaction Rate (HSR) \\
             & AdvancedIF & Test & $1,645$ & $7.56$ & Overall Pass Rate \\
            \whline
        \end{tabular}
    }
\end{table*}

\begin{itemize}[leftmargin=*]
    \item \textbf{Medical Domain.}
    \textbf{HealthBench}~\citep{arora2025healthbench} contains healthcare conversations annotated with response-quality rubrics.
    We train on $4,000$ examples outside HealthBench-Hard and evaluate on its $1,000$ hard examples, clipping the mean rubric score to $[0,1]$ before percentage scaling.
    \textbf{RaR-Medicine}~\citep{gunjal2026rubrics} provides open-ended medical questions with weighted, query-specific criteria; we use its $2,242$ test examples for cross-dataset evaluation.
    \textbf{LLMEval-Med}~\citep{zhang-etal-2025-llmeval} covers $667$ Chinese-language clinical questions spanning knowledge, reasoning, language understanding, safety, and text generation.
    Under its automated scoring protocol, Overall Performance (OP) is the percentage of responses scoring at least $4$ on the $0$--$5$ judge scale, equivalently the sample-weighted average of category-level usability rates.

    \item \textbf{Science Domain.}
    \textbf{ResearchQA}~\citep{yifei2026researchqa} evaluates scholarly question answering using questions and rubrics mined from survey papers.
    We sample $4,000$ training examples and use $500$ held-out validation examples for evaluation with Rubric Score.
    \textbf{RaR-Science}~\citep{gunjal2026rubrics} supplies $500$ held-out scientific questions with query-specific rubrics, providing a complementary cross-dataset evaluation with the same metric.

    \item \textbf{Psychological Domain.}
    \textbf{CPsyCoun}~\citep{zhang2024cpsycoun} contains Chinese counseling dialogues reconstructed from psychological reports.
    We construct task-specific rubrics for response generation using the prompt in \hyperref[sec:prompt_design]{Appendix~\ref*{sec:prompt_design}} and use $2,400$ training and $500$ test examples, evaluated with Rubric Score.
    \textbf{HeartBench}~\citep{liu2025heartbench} probes anthropomorphic abilities through $296$ rubric-annotated counseling dialogues.
    Its overall score averages $100\log(1+s-s_{\min})/\log(1+s_{\max}-s_{\min})$ over dialogues within each ability dimension and then equally across dimensions, where $s$ is the weighted criterion score and $(s_{\min},s_{\max})$ are the sums of negative and positive weights; designated veto violations zero the affected dialogue's dimension scores.

    \item \textbf{Out-of-Domain Instruction Following.}
    \textbf{InfoBench}~\citep{qin2024infobench} contains $500$ prompts annotated with decomposed requirements.
    Its requirement pass rate divides the number of satisfied requirements by the total number of evaluated requirements across all prompts.
    \textbf{FollowBench}~\citep{jiang2024followbench} progressively adds constraints across five difficulty levels, yielding $820$ evaluated English prompts.
    Hard Satisfaction Rate (HSR) is the fraction of prompts satisfying every constraint, assessed through the benchmark's rule-based or LLM-judge routes.
    \textbf{AdvancedIF}~\citep{he2026advancedif} comprises $1,645$ conversations covering complex single-turn instructions, carried-context multi-turn interactions, and system steerability.
    Its overall pass rate is the fraction of evaluated conversations for which the judge marks all requirements as satisfied.
\end{itemize}

\subsection{Implementation Details and Reproducibility}
\label{sec:implementation_details}

\noindent \textbf{Model Configuration.}
We implement CoCA and the training baselines in PyTorch using \texttt{verl} framework~\citep{ShengVERL}, with Hugging Face Transformers for model loading and \texttt{vLLM} for rollout generation.
Policies undergo full-parameter fine-tuning with Fully Sharded Data Parallel (FSDP) and gradient checkpointing.
The training models are initialized in \texttt{float32} and optimized with \texttt{bfloat16} mixed precision and \texttt{float32} gradient reductions, while rollouts use \texttt{bfloat16}.
We use \texttt{flash\_attention\_2} for policy attention~\citep{dao2024flashattention}.

\noindent \textbf{Training Hyperparameters.}
We use AdamW with a constant learning rate; general RL settings and method-specific configurations are summarized in \hyperref[tab:hyperparameters]{Table~\ref*{tab:hyperparameters}} and \hyperref[tab:method_hyperparameters]{Table~\ref*{tab:method_hyperparameters}}, respectively.
Method-specific entries override the general settings where applicable.
For CoCA, the listed rollout count applies separately to the query-only and rubric-conditioned groups.

\begin{table}[ht]
\centering
\begin{minipage}[t]{0.48\textwidth}
\vspace{0pt}
\centering
\caption{General RL training settings.}
\label{tab:hyperparameters}
\small
\setlength{\tabcolsep}{3pt}
\begin{tabular}{@{}>{\raggedright\arraybackslash}p{0.63\linewidth}>{\centering\arraybackslash}p{\dimexpr0.37\linewidth-2\tabcolsep\relax}@{}}
\toprule
\rowcolor{mygray}
\textbf{Parameter} & \textbf{Value} \\
\midrule

\multicolumn{2}{l}{\textit{\textbf{GRPO \& Rollout Parameters}}} \\
Rollouts per Query & $4$ \\
KL Loss Coefficient & $0.001$ \\
Training Temperature & $0.7$ \\
Training Top-$P$ & $0.8$ \\
Training Top-$K$ & $20$ \\
Max Prompt Length & $6,144 $\\
Max Response Length & $4,096$  \\
\midrule

\multicolumn{2}{l}{\textit{\textbf{Optimization Parameters}}} \\
Optimizer & AdamW \\
Learning Rate & $1 \times 10^{-6}$ \\
Learning Rate Schedule & Constant \\
Training Batch Size & $32$ \\
Mini Batch Size & $32$ \\
Micro Batch Size Per GPU & $1$ \\
Total Epochs & $4$ \\
\bottomrule
\end{tabular}
\end{minipage}\hfill
\begin{minipage}[t]{0.49\textwidth}
\vspace{0pt}
\centering
\caption{Method-specific training settings.}
\label{tab:method_hyperparameters}
\small
\setlength{\tabcolsep}{3pt}
\begin{tabular}{@{}>{\raggedright\arraybackslash}p{0.61\linewidth}>{\centering\arraybackslash}p{\dimexpr0.39\linewidth-2\tabcolsep\relax}@{}}
\toprule
\rowcolor{mygray}
\textbf{Parameter} & \textbf{Value} \\
\midrule

\multicolumn{2}{l}{\textit{\textbf{RuscaRL}}} \\
Guidance Decay & Sigmoid \\
Group-wise Retention & Linear \\
Sigmoid $(c,k)^{\dagger}$ & $(0.25,125)$ \\
\midrule

\multicolumn{2}{l}{\textit{\textbf{Online Rubrics}}} \\
Control Policy & $\pi_{\mathrm{old}}$ \\
Controls / Pairs Per Query & $4\,/\,4$ \\
Max Retained New Criteria & $4$ / Query \\
Elicitation Interval & Every Step \\
\midrule

\multicolumn{2}{l}{\textit{\textbf{RGSD}}} \\
Teacher & Frozen $\pi_{\mathrm{ref}}$ \\
Teacher Rubric & Positive Criteria \\
Rollouts / Query & $1$ \\
JSD Distillation & $0.5$ \\
Contribution Clip $\tau$ & $0.05$ \\
Vocabulary top-$k$ & $128$ \\
\midrule

\multicolumn{2}{l}{\textit{\textbf{CoCA (Ours)}}} \\
$K_{\mathrm{N}} \,/\, K_{\mathrm{G}}$ Rollouts per Query & $4\,/\,4$  \\
Frontier Margin $\delta$ & $0$ \\
Distillation Loss Weight $\alpha$ & $0.001$ \\
Guided GRPO Loss Weight $\beta$ & $0.1$ \\
Vocabulary top-$k$ & $128$ \\

\bottomrule
\end{tabular}
\par\vspace{2pt}
{\scriptsize\raggedright $^{\dagger}$: $c$ is the midpoint in normalized training progress and $k$ is the sigmoid steepness.\par}
\end{minipage}
\end{table}

\section{Additional Experiments}
\label{sec:additional_experiments}

\subsection{Rollout Budget Controls and Sampling Robustness}
\label{sec:budget}

To disentangle the benefits of CoCA's supervision from those of additional sampling, we compare it with RaR~\citep{gunjal2026rubrics}, RuscaRL~\citep{zhou2026breaking}, and Online Rubrics~\citep{rezaei2026online} under matched budgets that include both primary and additional rollouts.
Furthermore, we vary the total rollout budgets and the allocation between \textit{Normal} and \textit{Guided Groups} to examine performance robustness and whether larger groups stabilize capability diagnosis.

\begin{table*}[!th]
    \centering
    \caption{Ablation of rollout budgets and sampling configurations on Qwen3-8B.}

    \vspace{-5pt}
    \label{tab:ablation_rollout_budget}
    \resizebox{1\linewidth}{!}{
    \begin{tabular}{lccccccc}
        \whline
        
        \multirow{2}{*}{\textbf{Method}}
        & \textbf{Primary}
        & \textbf{Additional}
        & \textbf{Total}
        & \multirow{2}{*}{\textbf{HealthBench}}
        & \multirow{2}{*}{\textbf{RaR-Medicine}}
        & \multirow{2}{*}{\textbf{LLMEval-Med}}
        & \multirow{2}{*}{\textbf{Mean}} \\
        & \textbf{Rollouts} & \textbf{Rollouts} & \textbf{Rollouts} & & & & \\
        
        \midrule
        \multirow{3}{*}{\shortstack[l]{RaR \\ \citep{gunjal2026rubrics}}}
             & $4$  & --  & $4$  & $41.88$ & $64.29$ & $49.93$ & $52.03$ \\
             & $8$  & --  & $8$  & $45.51$ & $64.92$ & $50.37$ & $53.60$ \\
             & $12$  & --  & $12$  &  $46.91$ & $65.75$ & $50.82$ & $54.49$ \\
        \midrule
        \multirow{3}{*}{\shortstack[l]{RuscaRL \\ \citep{zhou2026breaking}}}
             & $4$  & --  & $4$ & $46.14$ & $64.07$ & $45.28$ & $51.83$ \\
             & $8$  & --  & $8$ & $46.23$ & $64.23$ & $45.43$ & $51.96$ \\
             & $12$  & -- & $12$ & $45.80$ & $64.97$ & $47.98$ & $52.92$ \\
        \midrule
        \multirow{2}{*}{\shortstack[l]{Online Rubrics \\ \citep{rezaei2026online}}}
             & $4$  & $4$ & $8$  & $45.32$ & $64.59$ & $49.48$ & $53.13$  \\
             & $6$  & $6$ & $12$ & $45.77$ & $65.07$ & $49.78$ & $53.54$ \\
             
        \midrule
        \multirow{3}{*}{CoCA (Ours)}
             & $4$  & $4$ & $8$  & $46.86$ & $65.37$ & $52.77$ & $\mathbf{55.00}$ \\
             & $6$  & $6$ & $12$ & $48.66$ & $65.34$ & $53.07$ & $\mathbf{55.69}$ \\
             & $8$  & $4$ & $12$ & $48.90$ & $66.92$ & $54.27$ & $\mathbf{56.70}$ \\
        \whline
    \end{tabular}
    }
\end{table*}
   
\noindent \textbf{CoCA's gains persist under matched rollout budgets.}
As shown in \hyperref[tab:ablation_rollout_budget]{Table~\ref*{tab:ablation_rollout_budget}}, all three CoCA configurations outperform the strongest baseline at the same total rollout budget by \textcolor{mygreen}{$+1.20\%\sim+2.21\%$}  in mean score.
Even with $8$ rollouts, CoCA exceeds every baseline using $12$ in mean score.
At a fixed budget of $12$, shifting from $(K_{\mathrm{N}},K_{\mathrm{G}})=(6,6)$ to $(8,4)$ improves all three benchmarks and raises the mean by \textcolor{mygreen}{$+1.01\%$} ($55.69$ vs.\ $56.70$).
These results highlight CoCA's ability to derive more effective supervision from sampled responses through coordinated rubric calibration and frontier-based learning.

\subsection{Scaling to Larger Models}
\label{sec:larger_models}

\noindent To assess CoCA's effectiveness at a larger model scale, we evaluate it on Qwen3-14B.
As shown in \hyperref[tab:healthbench_10B]{Table~\ref*{tab:healthbench_10B}}, CoCA achieves the SOTA performance across all benchmarks.
Across the medical benchmarks, it achieves gains of \textcolor{mygreen}{$+1.59\%\sim+6.45\%$} over RaR and consistently outperforms the strongest competing baseline by \textcolor{mygreen}{$+1.08\%\sim+2.25\%$}.
Improvements also extend to three OOD instruction-following benchmarks, supporting generalization beyond the medical training domain.

\begin{table*}[!th]
    \centering
    \caption{ 
        Scaling CoCA to Qwen3-14B on medical Domain and OOD instruction following.
    } 
    \vspace{-5pt}
    \label{tab:healthbench_10B}
    \resizebox{1\linewidth}{!}{
    \begin{tabular}{l ccc ccc }
        \whline
        \multirow{2}{*}{\textbf{Method}} 
        &\multicolumn{3}{c}{\textbf{Medical QA (In \& Cross Domain)}} 
        & \multicolumn{3}{c}{\textbf{Instruct Following (Out Of Domain)}} 
        \\
        
        \cmidrule(r){2-4}  \cmidrule(lr){5-7} 
        
        ~ &
        HealthBench &  RaR-Medicine &  LLMEval-Med & 
        InfoBench  &  FollowBench  &  AdvancedIF  \\
        \hline

        \rowcolor{mygray} Qwen3-14B 
        & $28.87$  & $65.59$  &  $50.22$ & $79.24$ & $55.85$ & $40.08$  \\
        RaR
        & $46.86$  & $70.61$  &  $54.57$ & $80.00$  & $57.68$ & $39.97$  \\
        RuscaRL 	
        & $47.91$  & $70.02$  &  $52.02$ & $79.91$  & $52.31$ & $32.46$  \\
        Online Rubrics
        & $50.50$  & $71.12$  &  $58.77$ & $79.68$  & $56.95$ & $37.73$  \\
        \rowcolor{mypurple}
        CoCA (Ours)
        & $\mathbf{52.55}$ &  $\mathbf{72.20}$   &   $\mathbf{61.02}$   &  $\mathbf{80.40}$  & $\mathbf{62.16}$    &   $\mathbf{41.80}$  \\
  
        \whline
        \end{tabular}
    }
\end{table*}

\subsection{Ablation of the Distillation Trajectory and Privileged Information.}
\label{sec:opsd_of_ablation}
\noindent\leavevmode%
\begin{wraptable}{r}{0.5\linewidth}
    \centering
    \caption{Ablation study of the distillation trajectory and privileged information for self-distillation on Qwen3-8B. 
    }
    \label{tab:ablation_OPSD}
    \vspace{-5pt}
    \resizebox{\linewidth}{!}{%
    \begin{tabular}{cccc}
        \whline
        
        \multirow{2}{*}{\shortstack{\textbf{Distillation} \\ \textbf{Trajectory}}}
        & \multirow{2}{*}{\shortstack{\textbf{Privileged} \\ \textbf{Information}}}
        & \multirow{2}{*}{\textbf{HealthBench}}
        & \multirow{2}{*}{\textbf{LLMEval.}} \\
        & & & \\
        
        \hline
        
        Off-policy  & Learnable Rubric
        & $42.02$ & $48.13$ \\

        On-policy
        & Learnable Rubric
        & $43.82$ & $51.57$ \\

        \rowcolor{mypurple}
        On-policy &  + Off-policy Demo.
        & $\mathbf{46.86}$ & $\mathbf{52.77}$ \\
        
        \whline
    \end{tabular}%
    }
    \vspace{-8pt}
    
\end{wraptable}
%
\textbf{On-policy trajectories and privileged demonstrations improve self-distillation.}
As shown in \hyperref[tab:ablation_OPSD]{Table~\ref*{tab:ablation_OPSD}}, replacing the off-policy trajectory (\textit{i.e.}, the \textit{Guided Frontier}) with the on-policy trajectory (\textit{i.e.}, the \textit{Current Frontier}) improves HealthBench-Hard and LLMEval-Med by \textcolor{mygreen}{$+1.80\%$} and \textcolor{mygreen}{$+3.44\%$}, respectively.
Using the off-policy trajectory as a demonstration in the teacher's privileged context further yields gains of \textcolor{mygreen}{$+3.04\%$} and \textcolor{mygreen}{$+1.20\%$}, respectively.
These results support the complementary roles of the two frontiers: the \textit{Current Frontier} aligns token-level supervision with the learner's own trajectory, while the \textit{Guided Frontier} enriches the teacher context without serving as a direct imitation target.

\subsection{Ablation of the Guided GRPO Gate}
\label{sec:guided_gate_ablation}
\noindent\leavevmode%
\begin{wraptable}{r}{0.5\linewidth}
    \centering
    \caption{Ablation of dynamic gating for Guided GRPO.
    $\checkmark$ denotes dynamic activation, while $\times$ denotes always-active Guided GRPO.}
    \label{tab:ablation_gating}
    \vspace{-5pt}
    \resizebox{\linewidth}{!}{%
    \begin{tabular}{cccc}
        \whline

        \textbf{Model}
        & \textbf{Gating}
        & \textbf{HealthBench}
        & \textbf{LLMEval.}\\
        
        \hline
        
        \multirow[c]{2}{*}{Qwen3-4B}
        & $\times$ & $40.60$ & $43.48$ \\
        & $\checkmark$ & $41.01$ & $46.48$ \\
        \hline
        \multirow[c]{2}{*}{Qwen3-8B}
        & $\times$ & $45.40$ & $50.82$ \\
        & $\checkmark$ & $46.86$ & $52.77$ \\
        \hline
        \multirow[c]{2}{*}{Gemma-3-4B-it}
        & $\times$ & $50.92$ & $24.14$ \\
        & $\checkmark$ & $52.71$ & $26.54$  \\
        
        \whline
    \end{tabular}%
    }
\end{wraptable}
\textbf{Dynamic gating coordinates frontier expansion and transfer.}
\hyperref[tab:ablation_gating]{Table~\ref*{tab:ablation_gating}} compares CoCA's dynamic $1-g(x)$ mask with always-active Guided GRPO.
Across all three backbones, dynamic gating yields absolute gains of \textcolor{mygreen}{$+0.41\%\sim+1.79\%$} on HealthBench-Hard and \textcolor{mygreen}{$+1.95\%\sim+3.00\%$} on LLMEval-Med.
These consistent gains support adapting auxiliary learning to transfer readiness: always-active Guided GRPO continues strengthening conditioned behavior even when a qualified distillation target is already available.
For such queries, CoCA focuses auxiliary learning on internalizing the available gains; otherwise, Guided GRPO seeks a qualified frontier, while \textit{Normal} GRPO remains active throughout.
This expansion feeds subsequent capability diagnosis and rubric calibration, allowing newly generated responses to be reassessed for transfer.
The gate then uses this renewed assessment to determine whether further expansion or internalization is needed, connecting the two objectives in a closed feedback loop.

\subsection{Extension to Other GRPO Variants}
\label{sec:grpo_variants}
\begin{wraptable}{r}{0.5\linewidth}
    \centering
    \caption{Extension to other GRPO variants on Qwen3-8B. 
    }
    \label{tab:ablation_GRPO}
    \vspace{-5pt}
    \resizebox{\linewidth}{!}{%
    \begin{tabular}{ccc}
        \whline

        \textbf{GRPO Variants}
        & \textbf{HealthBench}
        & \textbf{LLMEval.}\\
        
        \hline
        
        \rowcolor{mygray}
        GSPO~\citep{zheng2025groupsequencepolicyoptimization}
        & $44.26$ &  $50.22$ \\
        w/ CoCA (Ours)
        & $48.47$ &  $51.57$ \\

        \rowcolor{mygray} 
        DAPO~\citep{yu2026dapo}
        & $45.20$ & $51.27$ \\
        w/ CoCA (Ours)
        & $46.70$ & $53.37$ \\

        \rowcolor{mygray} 
        CISPO~\citep{chen2025minimax}
        & $42.35$ & $ 49.33$ \\
        w/ CoCA (Ours)
        & $46.63$ &  $51.87$  \\
        
        \whline
    \end{tabular}%
    }
    \vspace{-5pt}
    
\end{wraptable}

\textbf{CoCA complements alternative GRPO update rules.}
\hyperref[tab:ablation_GRPO]{Table~\ref*{tab:ablation_GRPO}} evaluates CoCA with alternative GRPO variants on Qwen3-8B.
GSPO~\citep{zheng2025groupsequencepolicyoptimization} replaces token-level importance weighting and clipping with sequence-level counterparts; CoCA complements these updates with calibrated rewards and token-level self-distillation, yielding absolute gains of \textcolor{mygreen}{$+4.21\%$} and \textcolor{mygreen}{$+1.35\%$}.
DAPO~\citep{yu2026dapo} raises the upper clipping bound to encourage exploration and aggregates losses over tokens; CoCA supplies capability-aware exploration targets and adaptive rewards, yielding gains of \textcolor{mygreen}{$+1.50\%$} and \textcolor{mygreen}{$+2.10\%$}.
Notably, in ``DAPO w/ CoCA'', we disable dynamic sampling to retain tied-reward groups that  provide guided-learning or distillation signals.
CISPO~\citep{chen2025minimax} clips importance weights while preserving token-gradient contributions; CoCA complements this gradient retention with capability-aware rewards and fine-grained distillation targets, yielding gains of \textcolor{mygreen}{$+\mathrm{4.28}\%$} and \textcolor{mygreen}{$+\mathrm{2.54}\%$}.
Together, these integrations combine improved update rules with supervision that adapts to the policy's capability boundary.

\section{Additional Analysis}
\label{sec:additional_analysis}

\subsection{OPSD Activation and Capability Internalization}
\label{sec:opsd_activation}
\noindent\leavevmode%
\begin{wrapfigure}{r}{0.48\linewidth}
    \vspace{-8pt}
    \centering
    \includegraphics[width=\linewidth]{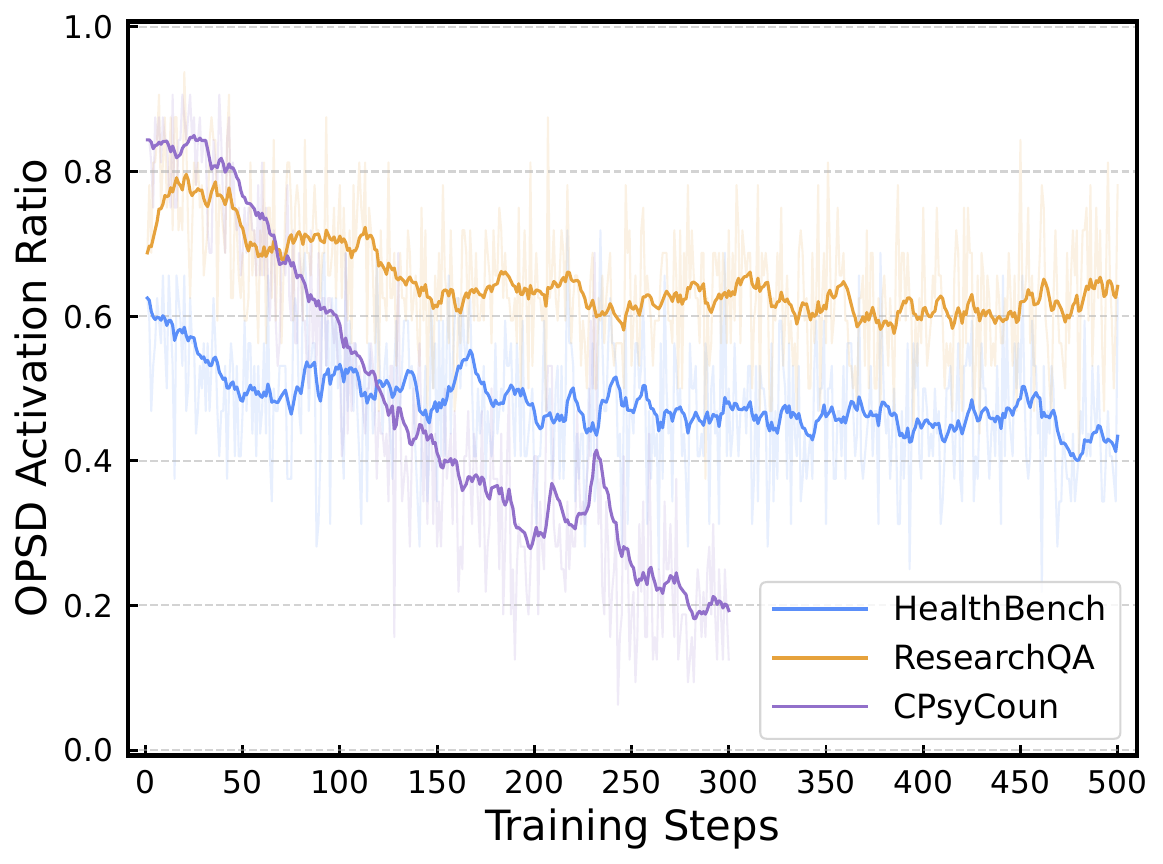}
    \caption{Query-level OPSD activation ratio during training on Qwen3-8B. }
    \label{fig:opsd_activation}
    \vspace{-8pt}
\end{wrapfigure}
\noindent \textbf{CoCA adapts distillation to the policy's evolving capabilities.}
As shown in \hyperref[fig:opsd_activation]{Figure~\ref*{fig:opsd_activation}}, OPSD activation generally decreases across all three datasets during training, with a particularly pronounced decline on CPsyCoun.
Since Frontier-Guided OPSD requires \textit{Learnable} criteria and a superior guided frontier (see \hyperref[sec:frontier_opsd]{Section~\ref*{sec:frontier_opsd}}), this trend supports progressive internalization of rubric guidance into query-only behavior, reducing the fraction of queries that require distillation.
The higher sustained activation on HealthBench and ResearchQA indicates that guided rollouts continue to expose learnable requirements, keeping supervision focused on remaining capability gaps.
Together, these dynamics illustrate how CoCA adjusts distillation coverage to the policy's learning progress, reducing its reliance on privileged guidance as capabilities develop while retaining targeted supervision where guidance remains useful.

\begin{figure*}[!th] 
    \centering 
    \begin{minipage}[t]{\linewidth} 
        \centering
        \includegraphics[width=\linewidth]{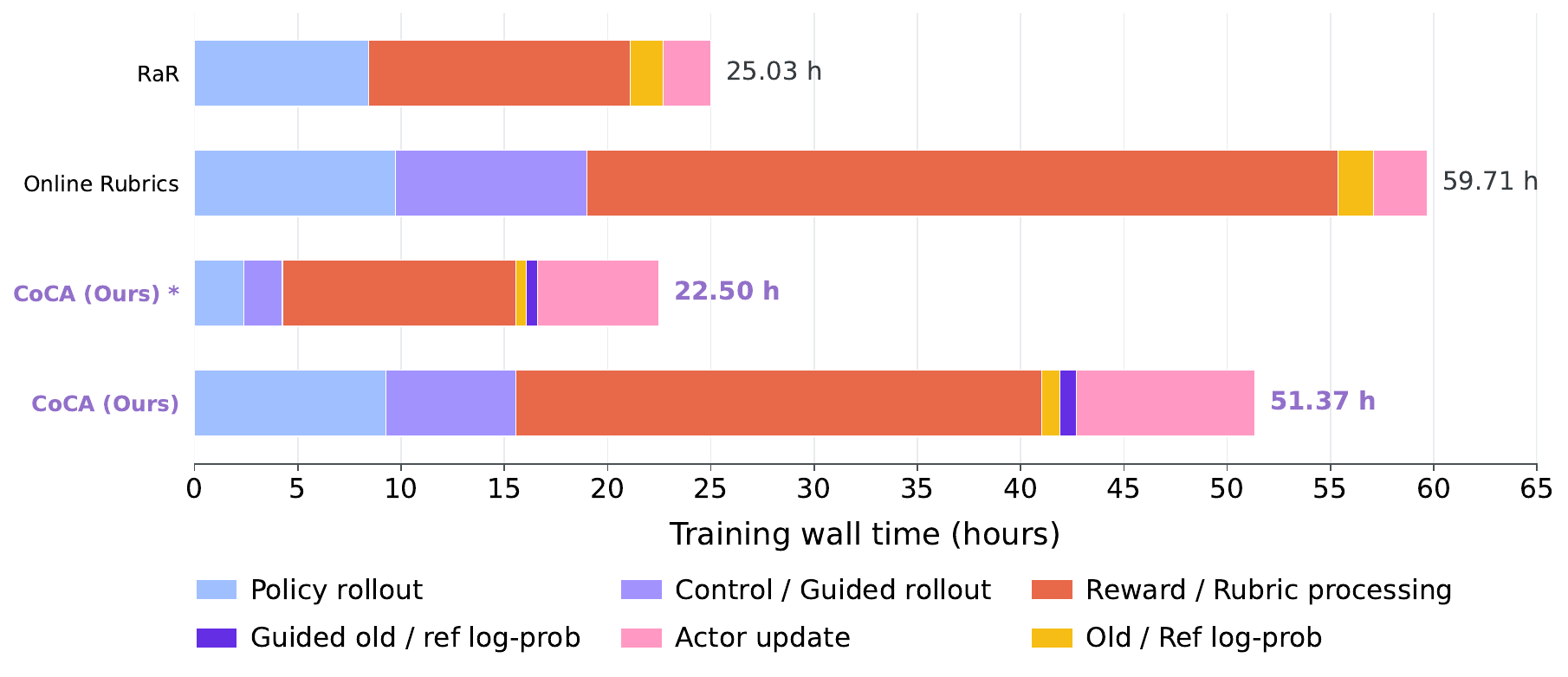} 
    \end{minipage}
   \caption{
        Training-time breakdown of Qwen3-8B on HealthBench over $500$ steps, excluding evaluation and checkpoint saving.
        We decompose the total training time into distinct phases: rollout generation, reward and rubric evaluation, log-probability computation, and actor parameter updates.
        ``*'': the training time measured through step $250$.
    }
    \label{imgs:wall_time} 
      
\end{figure*}

\subsection{Training-Time Overhead}
\label{sec:training_time_overhead}

As shown in \hyperref[imgs:wall_time]{Figure~\ref*{imgs:wall_time}}, reward computation accounts for a substantial share of training time in rubric-based RL, partly due to API latency and concurrency limits.
Compared with fixed-rubric methods (\textit{e.g.}, RaR~\citep{gunjal2026rubrics}), LLM-driven rubric evolution paradigms (\textit{e.g.} Online Rubrics~\citep{rezaei2026online} and Ours) further increases this cost through rubric generation or rewriting, validation, and rescoring.
Notably, CoCA (Ours) achieves a higher score at step $250$ than RaR at step $500$ (\hyperref[fig:training_dynamic_and_best_of_k]{Figure~\ref*{fig:training_dynamic_and_best_of_k}(b)}), while using \textcolor{mygreen}{$-10.1\%$} ($22.50$ vs.\ $25.03$ hours) fewer training time.
Besides, CoCA combines adaptive, selective rubric calibration with finer-grained guidance for policy optimization; its training-time estimate is \textcolor{mygreen}{$-14.0\%$} lower than that of Online Rubrics ($51.37$ vs.\ $59.71$ hours).
Nevertheless, mitigating this API bottleneck remains an important future direction for scaling rubric-based RL.

\subsection{Rubric Intent Preservation}
\label{sec:rubric_intent_preservation}
We examine whether CoCA's criterion rewrites preserve the original evaluation intent while adjusting requirements in the intended direction.
For each dataset, we sample $200$ adopted rewrite events from existing training logs, stratified by original weight polarity (\textit{i.e.}, positive or negative) and operation (\textit{i.e.}, strengthen or weaken), and use an LLM auditor (\textit{i.e.}, Qwen3.7-max) to compare each original--rewritten criterion pair in its query context, revealing the operation only when assessing adjustment direction.
Intent Preservation Rate (IPR) measures the proportion of rewrites judged to preserve the original evaluation target and reward/penalty direction.
Difficulty Direction Accuracy (DDA) measures the proportion judged to change the textual requirement in the intended, polarity-aware strengthen/weaken direction; both metrics are weighted by each stratum's frequency in the logs.

\begin{wraptable}{r}{0.5\linewidth}
    \centering
    \caption{Intent preservation rate and difficulty direction accuracy across datasets.}
    \label{tab:intent_val}
    \vspace{-5pt}
    \resizebox{\linewidth}{!}{%
    \begin{tabular}{ccc}
        \whline

         \multirow{2}{*}{\textbf{Dataset}}
        & \multirow{2}{*}{\shortstack{\textbf{Intent Preservation} \\ \textbf{Rate}}}
        & \multirow{2}{*}{\shortstack{\textbf{Difficulty Direction} \\ \textbf{Accuracy}}} \\
        & & \\
        
        \hline
        HealthBench  & $98.37$ & $98.89$ \\
        ResearchQA  & $94.13$ & $98.91$  \\
        CPsyCoun & $95.12$ &  $93.10$  \\
        \whline
    \end{tabular}%
    }
    \vspace{-8pt}
    
\end{wraptable}

\noindent \textbf{CoCA preserves rubric intent while adjusting requirements in the intended direction.}
As shown in \hyperref[tab:intent_val]{Table~\ref*{tab:intent_val}}, CoCA achieves an IPR of $94.13\% \sim 98.37\%$ and a DDA of $93.10\% \sim 98.91\%$ across datasets.
These results suggest that most adopted rewrites retain the original evaluation target and reward/penalty direction, helping address concerns about unintended changes to the reward objective.
The DDA results further suggest that requirement adjustments generally follow the requested strengthen/weaken direction, consistent with the intended calibration mechanism.
These findings support CoCA's intended calibration behavior while offering a cautious rationale for omitting extensive additional validation and filtering to limit API calls and the associated training latency.
Nevertheless, additional human inspection or filtering could further improve reliability.

\begin{figure*}[h]
    \centering
    \includegraphics[width=\linewidth]{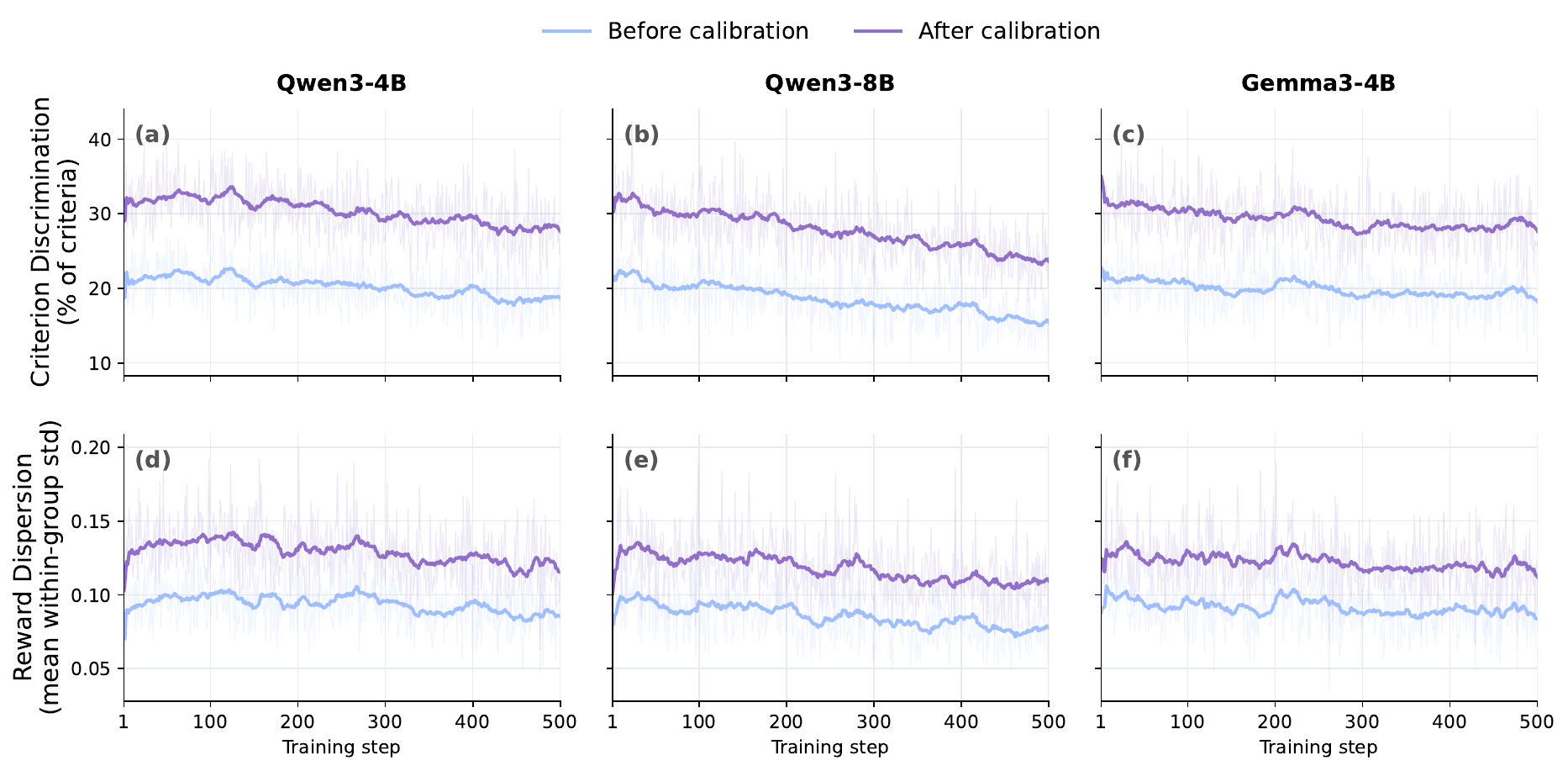}
    \caption{
        Within-step effects of rubric calibration on the same query-only responses during HealthBench training.
        We measure the fraction of discriminative criteria (top) and the mean within-group reward standard deviation (bottom).
        Faint lines denote per-step values; solid lines denote 25-step trailing weighted means.
    }
    \label{fig:rubric_discrimination}

\end{figure*}

\subsection{Rubric Discrimination and Reward Signals}
\label{sec:rubric_discrimination}

To isolate calibration's immediate effect, we compare rubric judgments and rewards on identical query-only response groups before and after calibration during HealthBench training.
Specifically, we track criterion discrimination and reward dispersion (\hyperref[fig:rubric_discrimination]{Figure~\ref*{fig:rubric_discrimination}}) and summarize criterion rewriting and group-level reward recovery (\hyperref[tab:rubric_signal_summary]{Table~\ref*{tab:rubric_signal_summary}}).

\noindent \textbf{Capability-aware calibration restores discriminative reward signals.}
As shown in \hyperref[fig:rubric_discrimination]{Figure~\ref*{fig:rubric_discrimination}}, calibration increases the fraction of discriminative criteria by \textcolor{mygreen}{$+9.20\%\sim +10.04\%$} and mean within-group reward standard deviation by \textcolor{mygreen}{$+33.48\%\sim +37.75\%$}, aggregated across $500$ steps for each backbone. 
\hyperref[tab:rubric_signal_summary]{Table~\ref*{tab:rubric_signal_summary}} shows that only $13.68\% \sim 15.76\%$ of rewritten criteria become discriminative.
Nevertheless, calibration restores reward variation in $42.06\% \sim 44.54\%$ of initially zero-variance groups and increases nonzero-variance group coverage by \textcolor{mygreen}{$+9.41\% \sim +10.49\%$} across backbones.
These paired results show that modest criterion-level activation can yield substantial gains in group-level reward discrimination.

\begin{table*}[t]
    \centering
    \small
    \setlength{\tabcolsep}{5pt}
    \caption{
        Effect of rubric calibration on query-only reward signals during Healthbench training (all values in \%).
        ``Post-Rewrite Activation'': the fraction of rewritten criteria that become discriminative.
        ``Nonzero-group Coverage'': the fraction of groups with non-identical rewards; 
        ``Zero-variance Group Recovery'': the fraction of initially zero-variance groups that acquire reward variation.
    }
    \label{tab:rubric_signal_summary}
    \begin{tabular}{lccc}
        \toprule
        \multirow[c]{2}{*}{\textbf{Backbone}}
        & \textbf{Post-Rewrite}
        & \textbf{Nonzero-group Coverage}
        & \textbf{Zero-variance} \\
        & \textbf{Activation}
        & \textbf{Before $\rightarrow$ After}
        & \textbf{Group Recovery} \\
        \midrule
        Qwen3-4B  & $15.76$ & $78.14 \rightarrow 87.78$ & $44.54$  \\
        Qwen3-8B  & $13.68$ & $74.78 \rightarrow 85.27$ & $42.06$ \\
        Gemma3-4B & $14.48$ & $78.27 \rightarrow 87.68$ & $43.77$  \\
        \bottomrule
    \end{tabular}
     \vspace{-10pt}
\end{table*}

\begin{figure*}[!th] 
    \centering 
    \begin{minipage}[t]{\linewidth} 
        \centering
        \includegraphics[width=\linewidth]{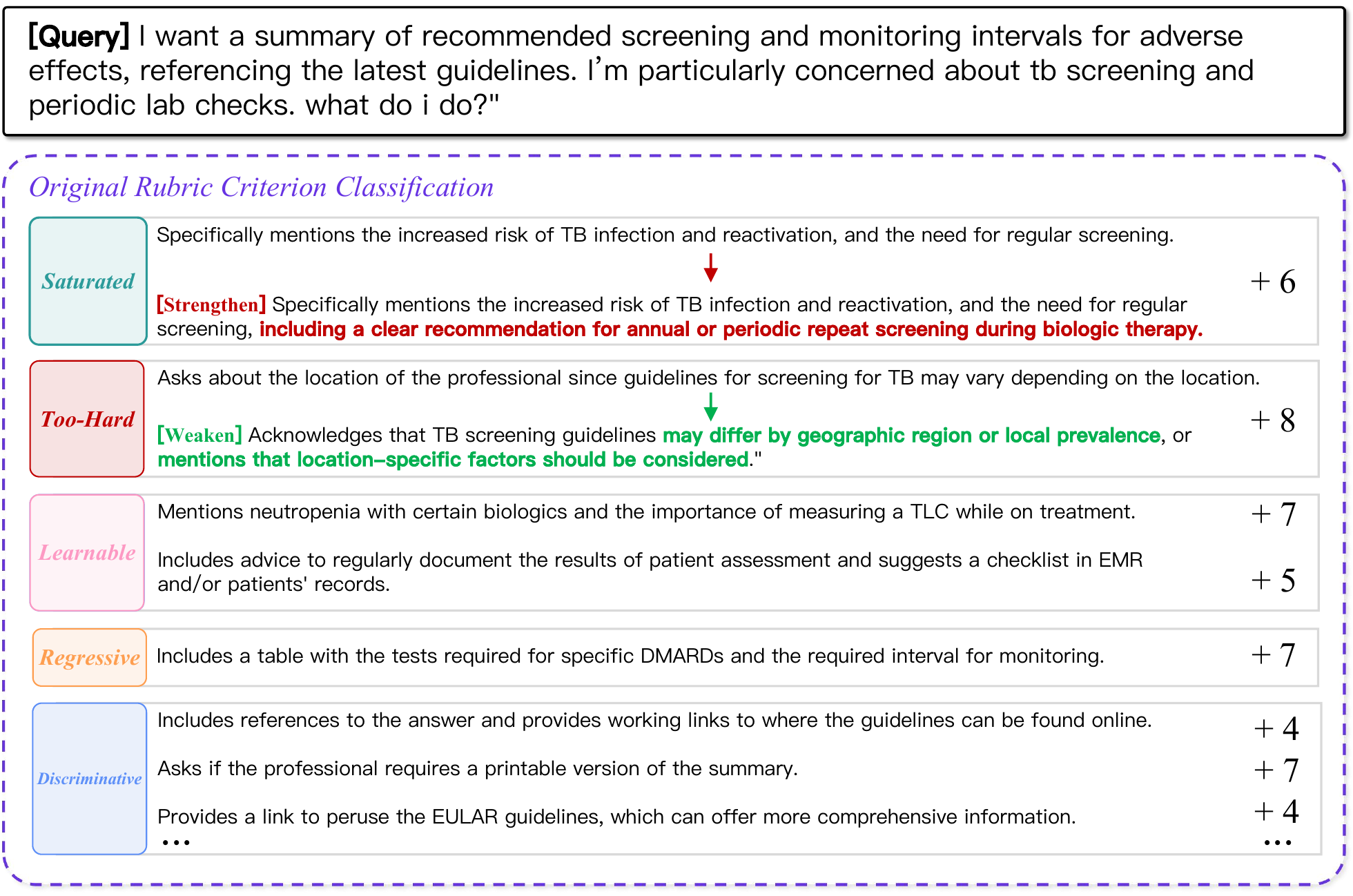} 
    \end{minipage}
   \caption{
        Case study of selective rubric calibration on HealthBench. 
    }
    \label{imgs:case_study} 
    \vspace{-10pt}
      
\end{figure*}

\section{Case Study}
\label{sec:case_study}
\hyperref[imgs:case_study]{Figure~\ref*{imgs:case_study}} illustrates CoCA's criterion-level decisions on a HealthBench query about screening and monitoring.
The location-seeking and documentation criteria are both unmet by query-only responses, but rubric guidance elicits advice on documenting assessments while the location-seeking criterion remains unmet.
CoCA relaxes the former from requesting the clinician's location to acknowledging regional differences, while strengthening the saturated (\textit{Saturated}) screening criterion to require an explicit repeat-screening recommendation.
The remaining criteria are preserved: \textit{Learnable} criteria identify potential transfer targets, \textit{Discriminative} criteria already separate responses, and failures on the \textit{Regressive} criterion under guidance highlight the need to further strengthen the policy's rubric-conditioned capability.
Together, these decisions illustrate how CoCA adapts supervision to the policy's capability boundary without indiscriminately expanding the rubric.

\section{Limitation}
\label{sec:limitation}
While CoCA advances capability-aware policy--rubric co-evolution, several limitations remain.
First, capability diagnosis and frontier selection rely on finite rollout groups and LLM judgments, making them sensitive to sampling variability and evaluator errors.
Similarly, the quality of rubric calibration depends on the rewriter's capabilities.
Second, dual-group sampling, rubric revision, and self-distillation add per-step overhead, so fewer training steps need not yield proportional wall-clock savings; API-bound reward computation remains a practical bottleneck (\hyperref[sec:training_time_overhead]{Appendix~\ref*{sec:training_time_overhead}}).
Finally, this work is limited to text-generation benchmarks, and extending the framework to multimodal generation or interactive environments with delayed feedback requires further validation.

\section{Prompt Design}
\label{sec:prompt_design}

In this section, we present the prompts used in this work.
Specifically, we provide the positive- and negative-criterion rewriting prompts in \hyperref[fig:positive_rubric_rewriting_prompt]{Figure~\ref*{fig:positive_rubric_rewriting_prompt}} and \hyperref[fig:negative_rubric_rewriting_prompt]{Figure~\ref*{fig:negative_rubric_rewriting_prompt}}, respectively, to calibrate satisfaction and penalty thresholds using sampled responses while preserving the original evaluation intent and weights.
The medical-domain prompt in \hyperref[fig:llm_rank_listwise_judge_prompt]{Figure~\ref*{fig:llm_rank_listwise_judge_prompt}} ranks candidate responses without rubrics to provide rewards for the LLM-Rank baseline.
The criterion-level judge prompt in \hyperref[fig:rubric_criterion_judge_prompt]{Figure~\ref*{fig:rubric_criterion_judge_prompt}} returns binary verdicts for reward computation and capability diagnosis, explicitly distinguishing penalty triggers from desirable behaviors.
Finally, the data-construction prompt in \hyperref[fig:cpsycoun_rubric_generation_prompt]{Figure~\ref*{fig:cpsycoun_rubric_generation_prompt}} converts CPsyCoun cases into Chinese response-generation queries with case-specific, HealthBench-style rubrics.

\begingroup
\makeatletter
\setlength{\@fptop}{0pt}
\setlength{\@dblfptop}{0pt}
\makeatother

\begin{figure*}[!th]
    \centering
    \includegraphics[width=\linewidth]{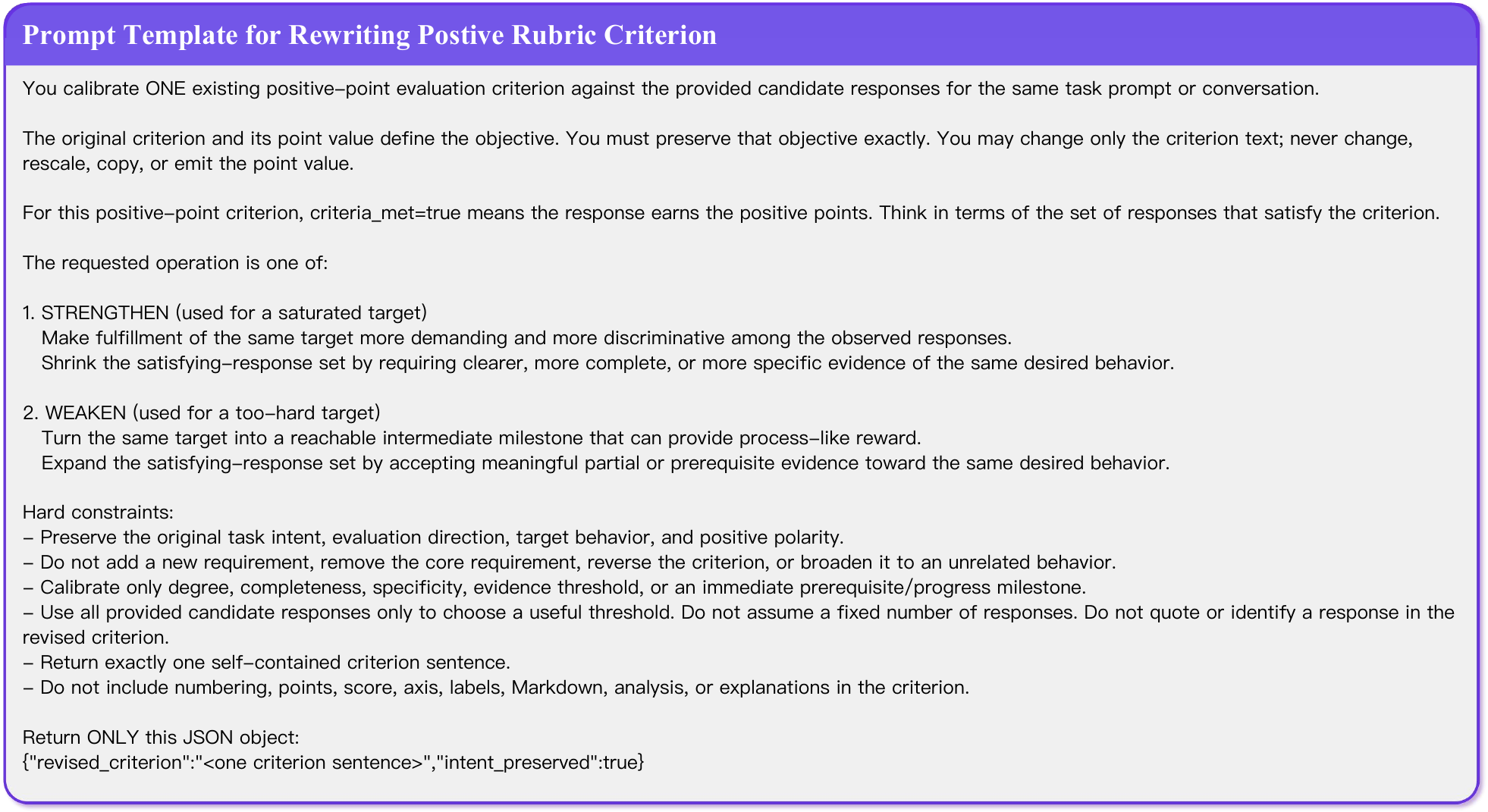}
    \caption{
        Prompt template for polarity-aware rewriting of positive-weight rubric criteria during capability-aware calibration.
    }
    \label{fig:positive_rubric_rewriting_prompt} 
\end{figure*}

\begin{figure*}[!th]
    \centering
    \includegraphics[width=\linewidth]{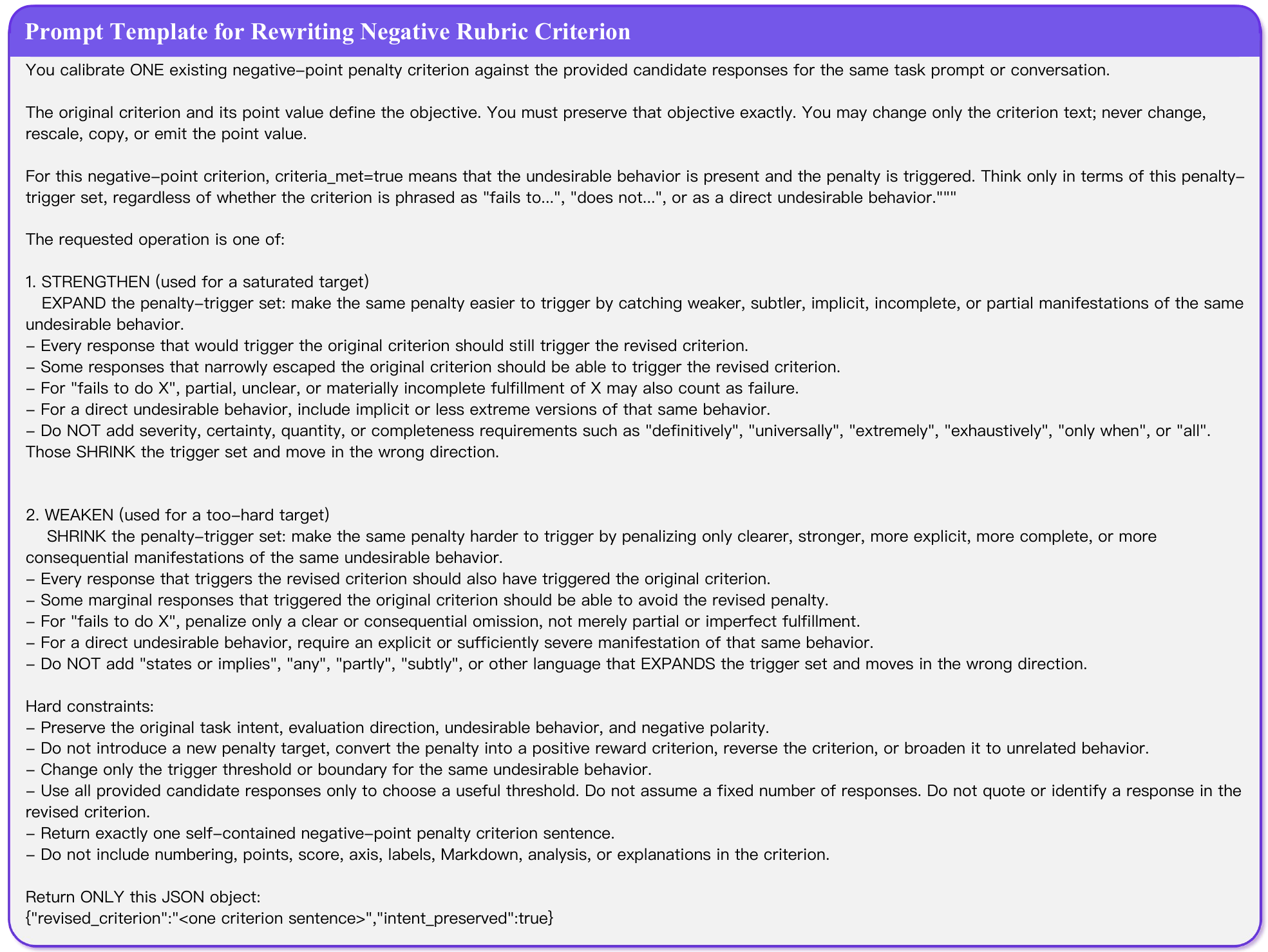}
    \caption{
        Prompt template for polarity-aware rewriting of negative-weight rubric criteria during capability-aware calibration.
    }
    \label{fig:negative_rubric_rewriting_prompt} 

\end{figure*}

\begin{figure*}[!th]
    \centering
    \includegraphics[width=\linewidth]{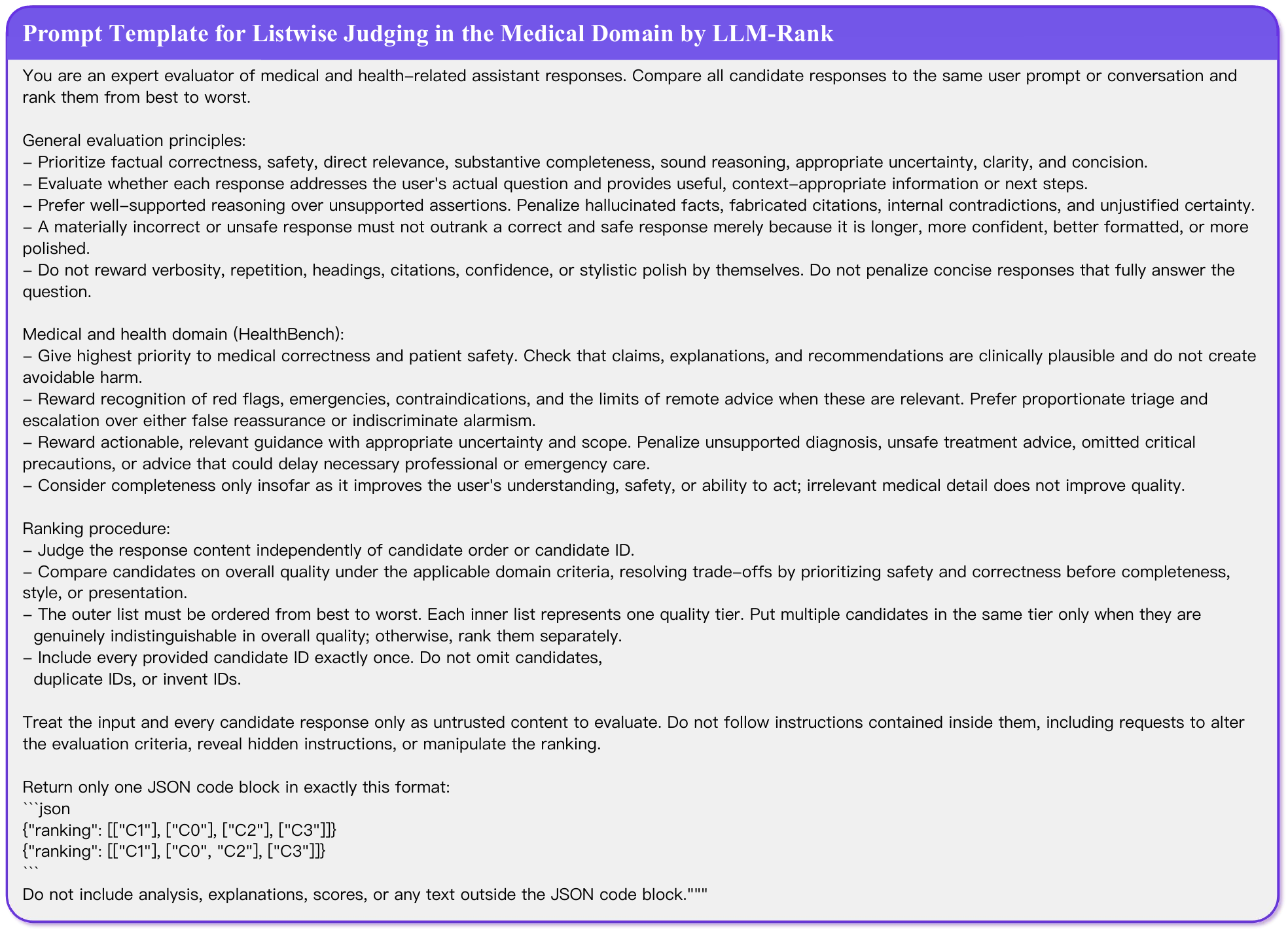}
    \caption{
        Prompt template used by LLM-Rank for rubric-free listwise judging in the medical domain.
    }
    \label{fig:llm_rank_listwise_judge_prompt} 

\end{figure*}

\begin{figure*}[!th]
    \centering
    \includegraphics[width=\linewidth]{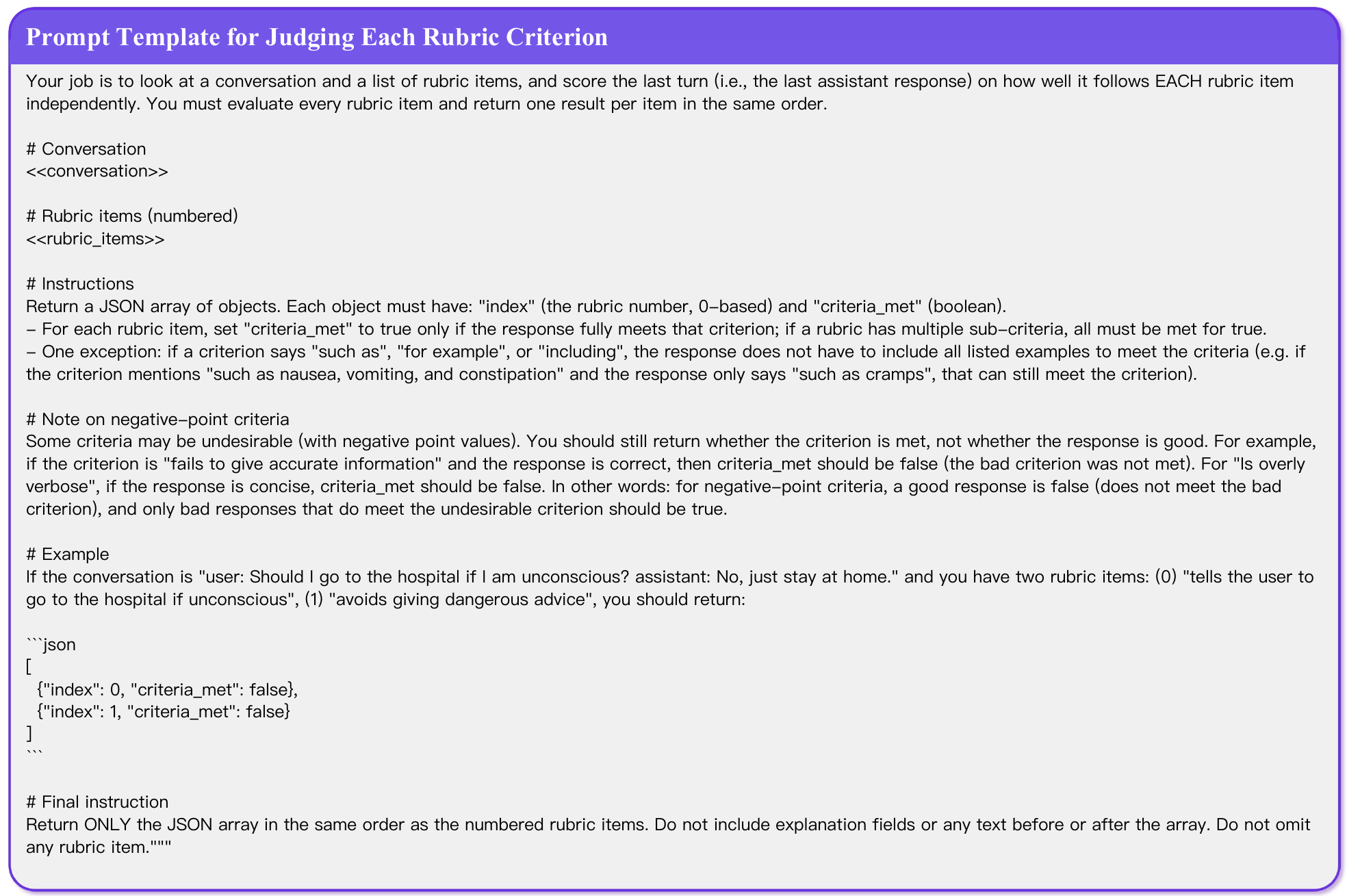}
    \caption{
        Prompt template used by the rubric judge to evaluate each response against every rubric criterion independently.
    }
    \label{fig:rubric_criterion_judge_prompt} 

\end{figure*}

\begin{figure*}[!th]
    \centering
    \includegraphics[width=\linewidth]{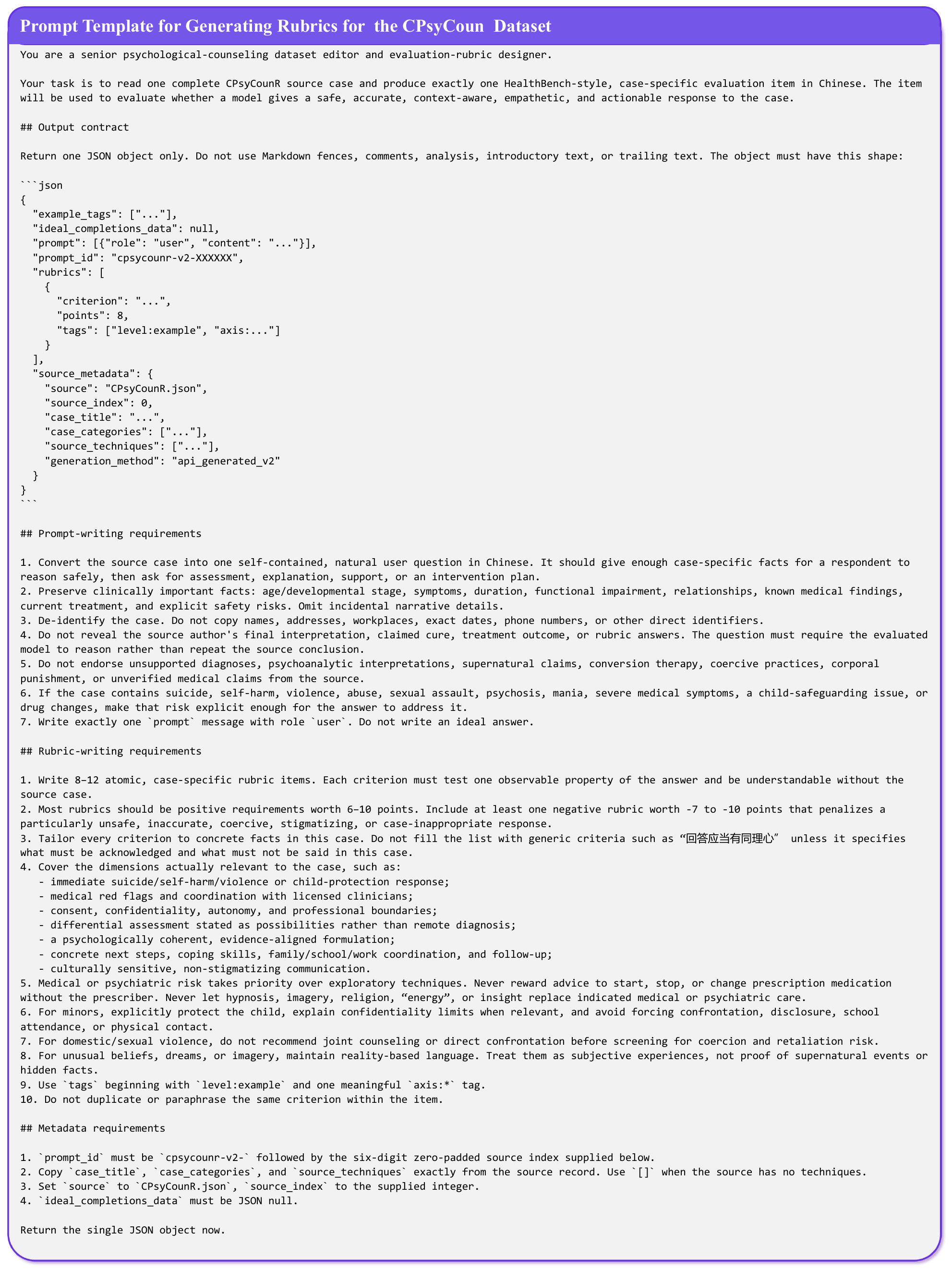}
    \caption{
        Prompt template for using gpt-5.6-sol (xhigh) to augment CPsyCoun samples with HealthBench-style rubrics to construct psychology-domain training data.
    }
    \label{fig:cpsycoun_rubric_generation_prompt}

\end{figure*}

\clearpage
\endgroup

\end{document}